\documentclass[11pt]{article}

\usepackage[final]{acl}

\usepackage{times}
\usepackage{latexsym}

\usepackage[T1]{fontenc}

\usepackage[utf8]{inputenc}

\usepackage{microtype}

\usepackage{inconsolata}

\usepackage{graphicx}

\usepackage{array}
\usepackage{lipsum}
\usepackage{multirow}
\usepackage{subcaption}
\usepackage{pifont}
\usepackage{skak}

\usepackage{graphicx}
\usepackage{booktabs}

\usepackage[accsupp]{axessibility}  

\providecommand{\inst}[1]{}

\usepackage{hyperref}
\usepackage{xurl} 
\usepackage{orcidlink}
\begin{document}

\title{Anthropogenic Regional Adaptation \\ in Multimodal Vision-Language Model} 

\author{
Samuel Cahyawijaya$^{\symqueen1,2}$,
Peerat Limkonchotiwat$^{\symqueen3,2}$\inst{},
Tack Hwa Wong$^{\symqueen4,2}$\inst{}, \\
\textbf{Hitesh Laxmichand Patel}$^{\symqueen5}$\inst{}, 
\textbf{Amit Agarwal}$^{\symqueen5}$\inst{}, 
\textbf{Manuel Antonio Rufino}$^{\symqueen6}$\inst{}, \\
\textbf{Carlos Rafael Catalan}$^{\symqueen6}$\inst{},
\textbf{Muhammad Reza Qorib}$^{\symrook7}$\inst{7}, 
\textbf{Vicky Feliren}$^{\symrook8}$\inst{}, \\
\textbf{Holy Lovenia}$^{\symrook2}$\inst{}, 
\textbf{Aye Hninn Khine}$^{\symrook9,2}$\inst{},
\textbf{Frederikus Hudi}$^{\symrook10,2}$\inst{}, \\
\textbf{David Anugraha}$^{\symrook11}$\inst{}, 
\textbf{Alham Fikri Aji}$^{\symrook12,2}$\inst{},
\textbf{Romrawin Chumpu}$^{\symrook13}$\inst{}, \\
\textbf{Viet-Thanh Pham}$^{\symknight14}$\inst{}, 
\textbf{Minghan Wang}$^{\symknight14}$\inst{},
\textbf{Mohamed Fazli Imam}$^{\symknight12}$\inst{}, \\
\textbf{Ruochen Zhang}$^{\symknight15,2}$\inst{},
\textbf{Joseph Marvin Imperial}$^{\symknight16,31}$\inst{},
\textbf{Khumaisa Nur'aini}$^{\symknight8}$\inst{}, \\
\textbf{Do Xuan Long}$^{\symknight13}$\inst{}, 
\textbf{Musa Izzanardi Wijanarko}$^{\symknight8}$\inst{},
\textbf{Joel Ruben Antony Moniz}$^{\symknight17}$\inst{}, \\
\textbf{Patrick Amadeus Irawan}$^{\symknight12}$\inst{},
\textbf{Hanif Muhammad Zhafran}$^{\symknight18}$\inst{}, 
\textbf{Isaiah Flores}$^{\symknight19}$\inst{}, \\
\textbf{Salsabila Zahirah Pranida}$^{\symknight12}$\inst{},
\textbf{Jun Kevin}$^{\symknight20}$\inst{}, 
\textbf{Jostin Jerico Rosal}$^{\symknight21}$\inst{}, \\
\textbf{Patricia Nicole Monderin}$^{\symknight6}$\inst{}, 
\textbf{Kun Kerdthaisong}$^{22}$,
\textbf{Ahmad Mustafid}$^{23}$, \\
\textbf{My Chiffon Nguyen}$^{2}$, 
\textbf{Natchapon Jongwiriyanurak}$^{24}$, 
\textbf{Siva Worajitwannakul}$^{16}$, \\
\textbf{Haochen Li}$^{25}$, 
\textbf{Adrian Xuan Wei Lim}$^{13}$, 
\textbf{Bin Wang}$^{26}$, \\
\textbf{Muhammad Ravi Shulthan Habibi}$^{27,2}$, 
\textbf{Lynnette Hui Xian Ng}$^{7}$, 
\textbf{Mithil Bangera}$^{28}$, \\
\textbf{Yeshil Bangera}$^{28}$, 
\textbf{Priyaranjan Pattnayak}$^{23}$, 
\textbf{Dun Li Chan}$^{29}$, \\
\textbf{Sherissa Caren Djuniwar}$^{30}$, 
\textbf{Cho Chan Myei Oo}$^{32}$,
\textbf{Hee Ming Shan}$^{12}$ \\
$^1$Cohere \quad
$^2$SEACrowd \quad
$^3$AI Singapore \quad \quad
$^4$Universiti Teknologi PETRONAS \quad \\
$^5$Oracle \quad
$^6$Samsung R\&D Institute Philippines \quad
$^7$Carnegie Mellon University \quad \\
$^8$Monash University, Indonesia \quad 
$^9$King Mongkut’s University of Technology Thonburi \quad \\
$^{10}$Nara Institute of Science and Technology \quad 
$^{11}$Stanford University \quad 
$^{12}$MBZUAI \quad  \\
$^{13}$National University of Singapore \quad 
$^{14}$Monash University, Australia \quad
$^{15}$Brown University \quad \\
$^{16}$University of Bath \quad
$^{17}$Mila - Quebec AI Institute \quad
$^{18}$Institut Teknologi Bandung \quad \\
$^{19}$Ateneo de Manila University \quad 
$^{20}$Universitas Pelita Harapan \quad \\
$^{21}$Seoul National University of Science and Technology \quad 
$^{22}$Thammasat University \quad \\
$^{23}$Independent \quad 
$^{24}$University College London \quad 
$^{25}$Nanyang Technological University \quad \\
$^{26}$MiroMind AI \quad 
$^{27}$University of Indonesia \quad 
$^{28}$University of New Haven \quad  \\
$^{29}$INTI International University and Colleges \quad 
$^{30}$Binus University \quad  \\
$^{31}$National University Philippines
$^{32}$ThoughtFull \\
\small{\textbf{$^{\symqueen}$Main Contributors}}\quad
\small{\textbf{$^{\symrook}$Major Contributors}}\quad
\small{\textbf{$^{\symknight}$Notable Contributors}}
}

\maketitle

\begin{abstract}

While the field of vision-language (VL) has achieved remarkable success in integrating visual and textual information across multiple languages and domains, there is no dedicated framework for assessing human-centric alignment in vision-language systems. 
We offer two contributions to address this gap. First, we introduce \textbf{Anthropogenic Regional Adaptation}: a novel paradigm that aims to optimize model relevance to specific regional contexts while ensuring the retention of global generalization capabilities. Second, we present a simple, but effective adaptation method named Geographical-generalization-made-easy (GG-EZ), which utilizes regional data filtering and model merging. 
Through comprehensive ablations on 3 VL architectures: large vision-language models (LVLM), text-to-image diffusion models, and vision-language embedding models, and a case study in Southeast Asia (SEA) regional adaptation, we demonstrate the importance of Anthropogenic Regional Adaptation and GG-EZ, showing 5-15\% gains in cultural relevance metrics across SEA while maintaining over 98\% of global performance and even occasionally surpassing it.
Our findings establish Anthropogenic Regional Alignment as a foundational paradigm towards applicability of multimodal VL architectures in diverse regions and demonstrate a simple-yet-effective baseline method that optimizes regional value alignment while preserving global generalization.\footnote{All datasets and models can be found \href{https://huggingface.co/collections/SEACrowd/sea-vl-phase-2-multimodal-vision-language-models-for-sea}{here}.}

\end{abstract}

\section{Introduction}
\label{sec:intro}

Representation alignment in underrepresented regional contexts of foundation models has been a longstanding problem in AI~\cite{kakwani-etal-2020-indicnlpsuite,winata-etal-2025-worldcuisines,anugraha2025m4}, limiting their effectiveness and applicability across diverse global populations. This challenge is particularly acute in under-developed and developing regions -- such as African~\cite{adnan2021masakha,wang2024afri}, Indian~\cite{kakwani-etal-2020-indicnlpsuite,verma2025milu}, Middle Eastern, Southeast Asian (SEA)~\cite{cahyawijaya-etal-2023-nusacrowd,lovenia-etal-2024-seacrowd,cahyawijaya-etal-2025-crowdsource,winata-etal-2023-nusax}, etc -- where existing resources exhibit severe underrepresentation. Building on this critical gap, we observe that existing vision-language (VL) AI solutions, despite their capabilities to generalize across different languages and domains, often exhibit cultural insensitivity~\cite{naous-etal-2024-beer,patel2025sweeval,cahyawijaya-etal-2025-high}, stereotypical outputs~\cite{liu-etal-2025-culturally,cahyawijaya-etal-2025-crowdsource}, and reduced task performance when deployed in underrepresented regions~\cite{kabra-etal-2023-multi,liu-etal-2024-multilingual,agarwal-etal-2025-aligning-llms}. This misalignment stems from training data dominated by certain regions with limited exposure to diverse contextual nuances prevalent in other regions~\cite{cahyawijaya-etal-2023-nusawrites,cahyawijaya2024llmeveryonerepresentingunderrepresented,adilazuarda-etal-2024-towards}, undermining their real-world applicability.

To bridge this divide, we propose Anthropogenic Regional Adaptation, a foundational paradigm designed to systematically evaluate human-centric alignment of VL models across regional contexts while preserving global generalization capabilities. Within this framework, we identify two primary model archetypes: (1) global models that achieve strong performance across broad geographic contexts but struggle with underrepresented regions, and (2) regionally specialized models that excel on locale-specific metrics, yet falter when confronted with broader global contexts. Neither archetype is ideal, creating a critical gap in the utilization of existing systems for targeted regional applications.

\begin{figure*}[!t]
    \centering
    \includegraphics[width=0.49\linewidth,height=0.32\linewidth]{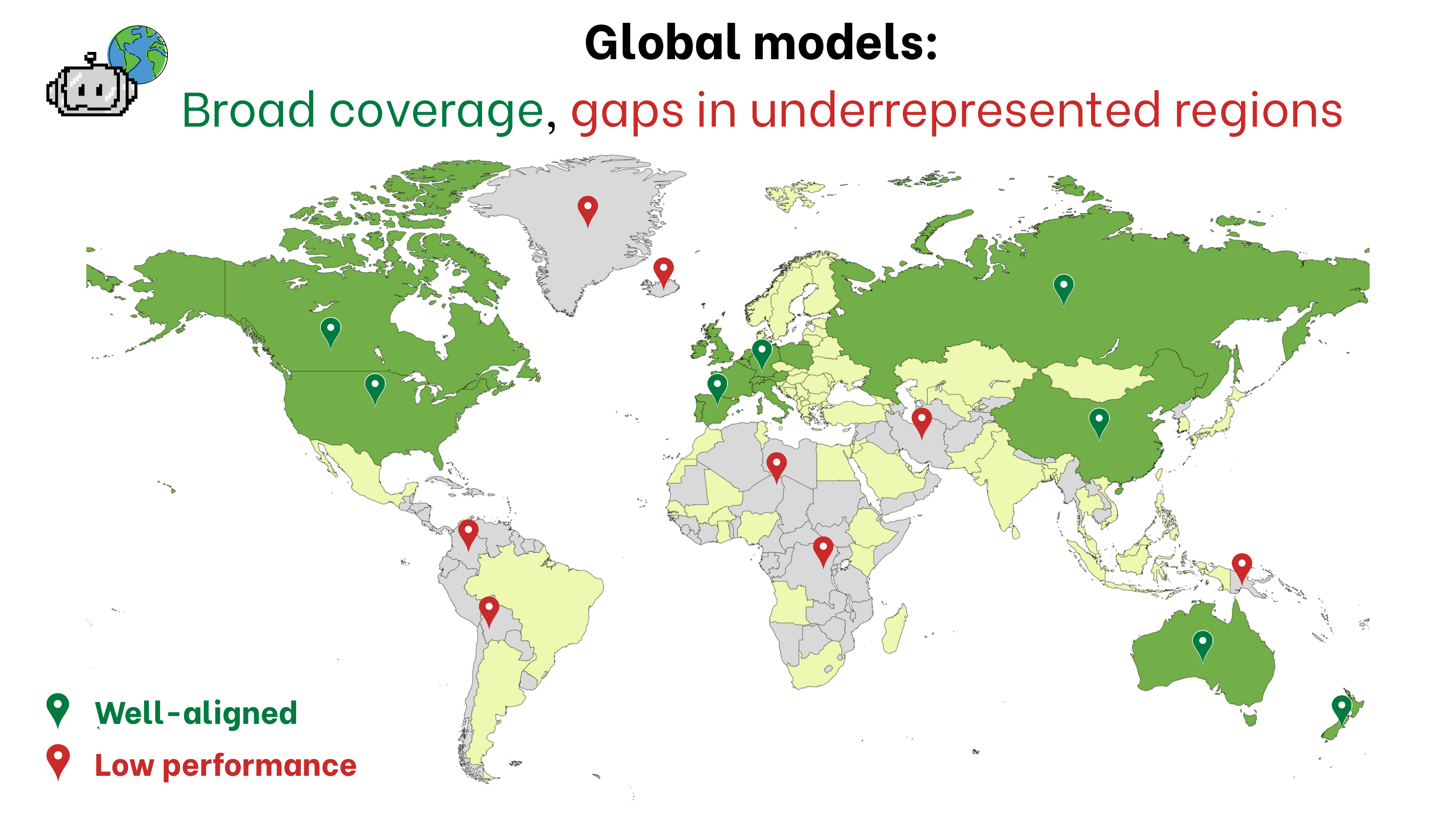}
    \includegraphics[width=0.49\linewidth,height=0.32\linewidth]{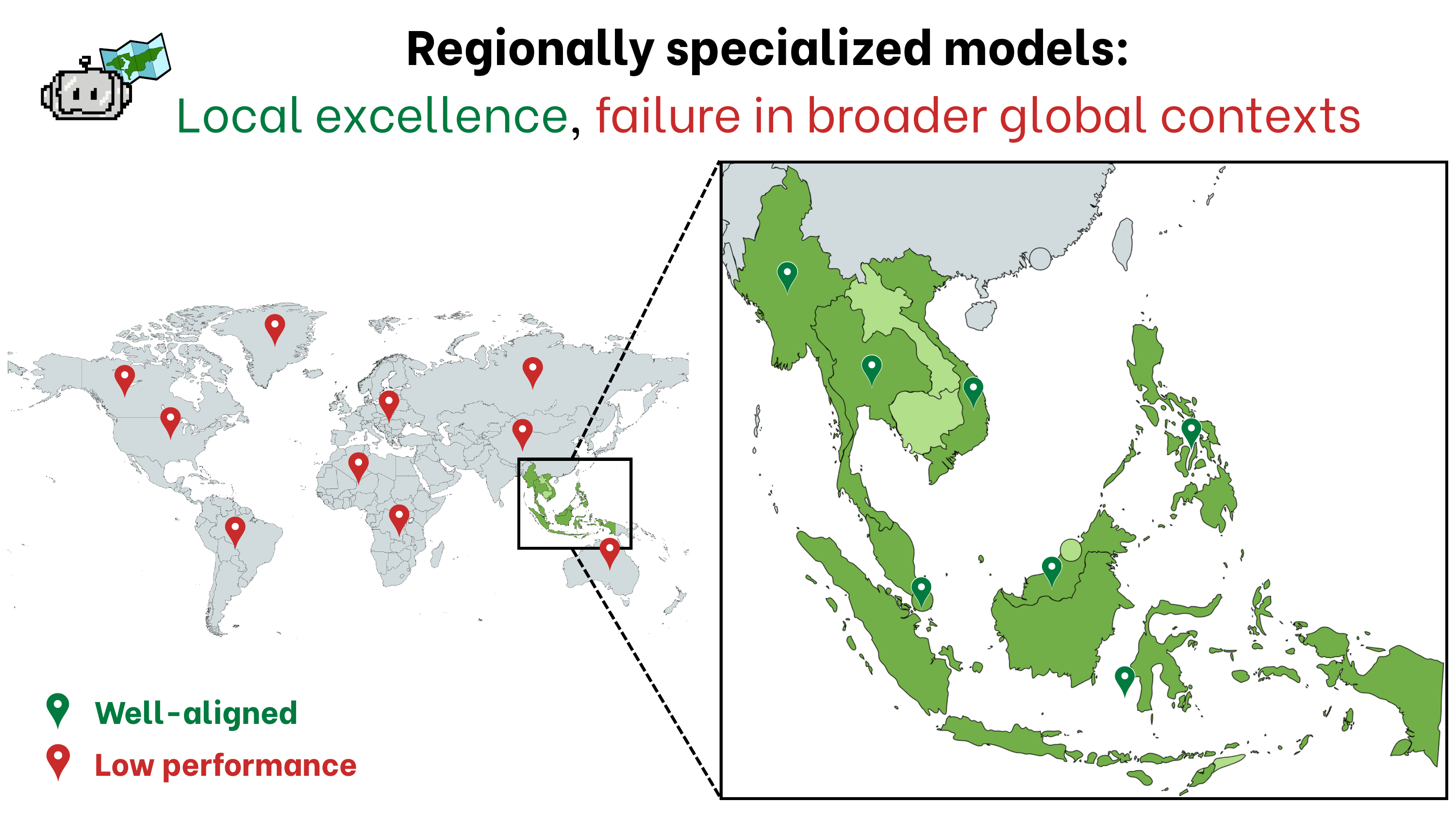}
    \caption{Through anthropogenic regional adaptation, we identify two primary model archetypes: \textbf{(left)} Global model with strong overall global performance, but struggle to represent certain regions appropriately, and \textbf{(right)} Regional-specific model that has a strong representation towards certain regions, but fall short on the global context.}
    \label{fig:highlight-figure}
\end{figure*}

Overcoming this limitation, we specifically designed a simple yet effective method, Geographical-generalization-made-easy (GG-EZ), which adapts an existing global model to a regional-specific context with minimal degradation on the global context. Inspired by recent advancements in training strategies of large language models (LLMs)~\cite{cohere2025commandaenterprisereadylarge,kocmi-etal-2025-command,zhang2025qwen3embedding,guo2025deepseekr1}, GG-EZ operationalizes regional adaptation through a two-level approach: (1) regional data filtering to curate culturally relevant training subsets, and (2) model merging to integrate region-specific adaptations without catastrophic forgetting of global knowledge.

We validate Anthropogenic Regional Adaptation and GG-EZ through rigorous experiments across three VL architectures -- large vision-language models (Gemma-3 27B), text-to-image diffusion models (SDXL), and vision-language embedding models (SigLIP-2) -- focusing on SEA as a case study. Our results demonstrate that GG-EZ achieves 5-15\% improvements in cultural relevance metrics, including cultural context accuracy and local visual context understanding, while retaining over 98\% of global performance on standard benchmarks. These findings establish Anthropogenic Regional Adaptation as a critical step toward equitable vision-language solutions and position GG-EZ as a practical baseline for anthropogenic regional adaptation in vision-language systems.

\section{Related Work}

Regional-specific models have been reported to outperform one-size-fits-all models on regional context \cite{nguyen-etal-2024-seallms}. Several datasets \cite{lovenia-etal-2024-seacrowd, cahyawijaya-etal-2025-crowdsource} and language models \cite{ng-etal-2025-sea, nguyen2026opensealgoodfastcheap} have been developed for the Southeast Asia (SEA) region.
However, in computer vision, work on regional adaptation has focused more on remote sensing~\cite{nyborg2022timematch,wang2025crossregional}, which results in limited understanding of a region's unique cultural and anthropomorphic characteristics. \citet{cahyawijaya-etal-2025-crowdsource} report that existing multilingual vision-language models such as MAYA-8B \cite{alam2024mayainstructionfinetunedmultilingual}, PaliGemma-2-10B \cite{steiner2024paligemma2}, Pangea-7B\cite{yue2024pangeafullyopenmultilingual}, Qwen2-VL-7B \cite{Qwen2-VL}, Gemma-3~\cite{gemma3_2025} fail in generating culturally-relevant responses when benchmarked against regional-specific cultural benchmarks such as SEAVQA \cite{urailertprasert-etal-2024-sea}, CVQA~\cite{mogrovejo2024cvqa} and World Cuisines \cite{winata-etal-2025-worldcuisines}. On the other hand, recent development of regional-specific vision-language models such as VIOLET~\cite{mohamed-etal-2023-violet}, Baseer~\cite{hennara2025baseervisionlanguagemodelarabic}, VARCO-VISION~\cite{ju2024varcovisionexpandingfrontierskorean}, and SEA-LION-VL~\cite{sea_lion_2024}, have strong performance on regional-specific benchmark, but fall short on a broader and more general global context.
Motivated by these limitations, we present a generic framework to systematically assess the gap of anthropogenic regional alignment in different models and introduce a simple-yet-effective approach, GG-EZ, to reduce the regional gap of existing multi-modal vision-language models, while maintaining its performance on the broader global contexts.\footnote{We provide comparison of our SEA-VLM models with other models in Appendix~\ref{sec:competitors-comparison}.}

\section{Anthropogenic Regional Adaptation}

\subsection{Overview}

Let $\mathcal{R}^{\text{glob}}$ denote the comprehensive spatial domain representing our global context of interest. Within this universal space, we establish a discrete partitioning into meaningful geographical or conceptual units through the definition of $\mathcal{R}$, a finite collection of $k$ distinct regions. Formally, $\mathcal{R} = \{R_1, R_2, \dots, R_k\}$ where each individual region $R_i$ constitutes a measurable subset of $\mathcal{R}^{\text{glob}}$. This decomposition enables systematic analysis of adaptation strategies across heterogeneous spatial contexts, recognizing that the world is rarely homogeneous in its characteristics or response patterns to anthropogenic influences.

Building upon this foundational regional partitioning, we introduce a critical distinction between regions based on strategic importance. The target region $\mathcal{R}^{\text{reg}}$ is defined as a specific, non-empty subset of $\mathcal{R}^{\text{glob}}$ representing our primary focus region where adaptation efforts should be maximized. Complementarily, the remaining regions are collectively designated as $\mathcal{R}^{\text{others}}$, defined through set difference as $\mathcal{R}^{\text{others}} = \mathcal{R}^{\text{glob}} \setminus \mathcal{R}^{\text{reg}}$. This ensures complete domain coverage while maintaining clear analytical separation between priority and non-priority regions, with each point in $\mathcal{R}^{\text{glob}}$ belonging to exactly one region.

Building upon this theoretical framework, as depicted in Figure~\ref{fig:highlight-figure}, existing vision-language models predominantly follow two archetypal approaches. Global models capture broad, universal patterns across $\mathcal{R}^{\text{glob}}$ but sacrifice nuanced regional specificity. Conversely, regional-specific models focus on particular $\mathcal{R}^{\text{reg}}$ subsets with high local performance but reduced global coherence. This dichotomy reflects a fundamental trade-off in model design between universal applicability and specialized excellence. Most existing models, commit to one of the paradigm: global optimization for comprehensive coverage or regional specialization for enhanced regional performance.

\subsection{Global-Regional Parity (GRP) Optimization}

For each region $R_i$, we have a set of evaluation metrics $E^{R_i} = \{e_1^{R_i}, e_2^{R_i}, \dots, e_m^{R_i}\}$. We define a quality metric $q_j^{R_i}$ for each evaluation metric $e_j^{R_i}$, and the collection of these quality metrics forms the set $Q^{R_i} = \{q_1^{R_i}, q_2^{R_i}, \dots, q_m^{R_i}\}$. This multi-dimensional approach captures the complexity of regional characteristics influencing adaptation outcomes. The evaluation framework extends to our partitioned structure, yielding aggregated evaluation sets $Q^{\mathcal{R}^{\text{glob}}}$ and $Q^{\mathcal{R}^{\text{reg}}}$ that consolidate metrics across regions while maintaining connections to individual characteristics. To balance competing objectives, we introduce a globalization factor $\alpha \in [0,1]$ that explicitly controls the trade-off between optimizing the target region and maintaining global generalization. The formal optimization objective is constructed as a weighted combination:
$$
\max_{\theta} \left[ (1 - \alpha) \cdot Q^{\mathcal{R}^{\text{reg}}} + \alpha \cdot Q^{\mathcal{R}^{\text{glob}}} \right]
$$

This formulation creates a single scalar objective that can be optimized using standard techniques, with $\alpha$ serves as a critical control mechanism that can be adjusted based on the specific focus and characteristics of each region. For regions with strong global connections or dependencies, $\alpha$ can be set closer to 1 to emphasize maintenance of $\mathcal{R}^{\text{glob}}$ conditions, while for regions where local impacts dominate, $\alpha$ can approach 0 to prioritize $\mathcal{R}^{\text{reg}}$ optimization. The solution achieves "best parity" between regions, representing the most favorable equilibrium where neither aspect is disproportionately favored according to our specified weighting scheme.

\section{Geographical Generalization Made Easy (GG-EZ)}

\begin{figure*}[!t]
    \centering
    \includegraphics[width=\linewidth, trim={0.5cm 11.5cm 0.5cm 0.5cm}, clip]{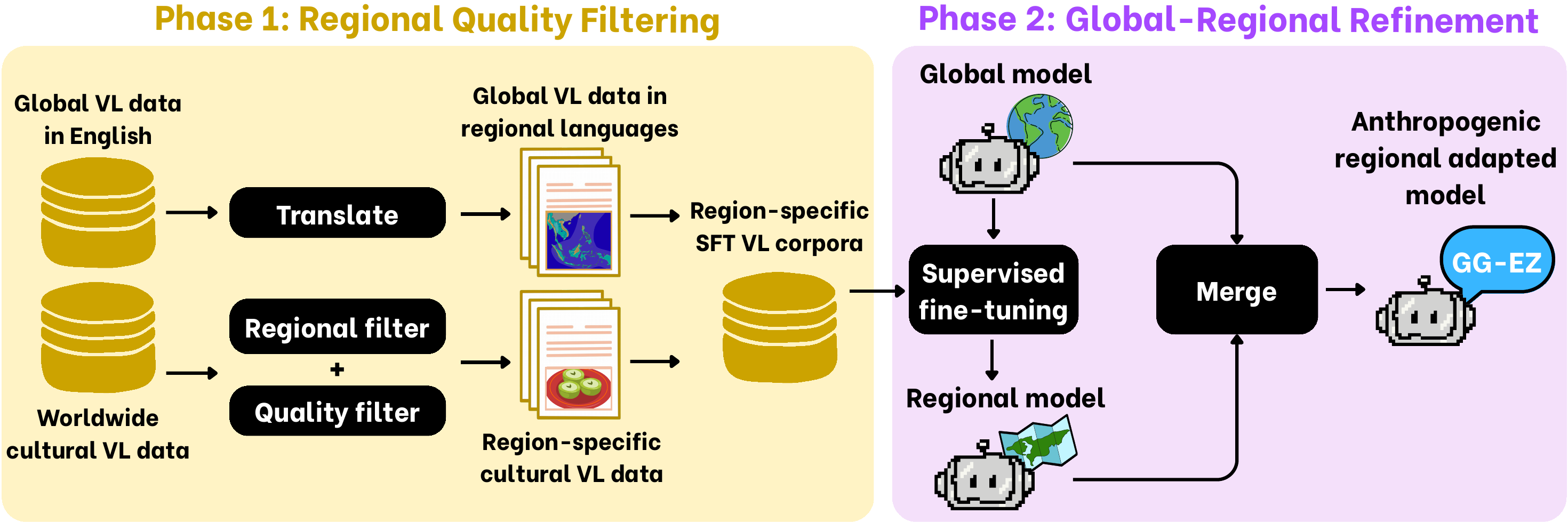}
    \caption{Overview of our Geographical-generalization-made-easy (GG-EZ) framework. Our framework consist of 3 constituents: (1) High-quality regional data filtering pipeline; (2) supervised fine-tuning to create a high quality regional-specific model; and (3) model merging to capture the best combination between regional-specific and global represention while also maintaining the generalization capabilities of the model.}
    \label{fig:gg-ez-method}
\end{figure*}

Geographical-generalization-made-easy (GG-EZ) is a general framework for regional adaptation of global multimodal models that achieves effective region-specific performance while preserving original generalization capabilities. As illustrated in Figure ~\ref{fig:gg-ez-method}, the core mechanism of GG-EZ consists of two phases: \textbf{regional quality filtering} and \textbf{global-regional refinement}. GG-EZ not only enables adaptation across local textual and imagery contexts while maintaining base model capabilities, but also better yet being architecture-agnostic, allowing effective regional adaptation across different vision-language architectures.

\subsection{Regional Quality Filtering}
The regional quality filtering phase refines the training dataset to focus on regionally relevant examples while preserving general knowledge. Let $D = \{D_{\text{reg}}, D_{\text{gen}}\}$ represent the complete dataset, where $D_{\text{reg}}$ contains region-specific examples and $D_{\text{gen}}$ contains general-domain examples. 

\textbf{Regional Filter ($F_{\text{rf}}$):} We define a boolean regional filter function that selects examples from the target region:
\[
F_{\text{rf}}(x, r) = \begin{cases} 
1 & \text{if } \text{region}(x) = r \\
0 & \text{otherwise}
\end{cases}
\]
where $r$ denotes the target region and $\text{region}(x)$ returns the geographical region associated with example $x$.

\textbf{Multilingual Reward ($F_{\text{rm}}$):} To further refine data filtering, we employ reward models that scores data quality and relevance on a continuous scale:
\[
F_{\text{rm}}(x) = \theta_{rwd}(x)
\]
where $\theta_{rwd}$ is the weight of the reward model and $\theta_{rwd}(x)$ measure the quality of the data $x$ using the reward model $\theta_{rwd}$. The filtered regional-specific dataset is then obtained by applying a threshold $\tau$:
\[
\small{D_{\text{filt}} = \{x \in D_{\text{reg}} \mid F_{\text{rf}}(x, r) = 1 \text{ and } F_{\text{rm}}(x) > \tau\}}
\]

\textbf{Language Translation Augmentation:} Beyond regional filtering of existing cultural-relevant corpora, a further data enrichment is done through translating English high-quality datasets to multiple target regional languages. Let $T$ represent the translation function mapping English to regional languages $L_r$:
\[
D_{\text{trans}} = \{T(x, \text{English} \rightarrow l) \mid x \in D_{\text{filt}}, l \in L_r\}
\]

\subsection{Global-Regional Refinement}

Using a pre-trained global model and the resulting corpora from the regional quality filtering phase, two steps of global-regional refinements are incorporated: 1) supervised fine-tuning, which turns the global model into a highly performant region-specific model, and 2) model merging to ensure the preservation of the original global generalization capabilities. We first trained the global model $\theta_{global}$ into a quality regional-specific model $\theta_{\text{reg}}$ using  $D_{\text{sft}} \subseteq \ D_{\text{filt}} \cup  D_{\text{trans}} $.
After supervised fine-tuning on $D_{\text{sft}}$, a linear model merging technique is applied to combine the global model with the region-specific model. The merged model parameters $\theta_{\text{merged}}$ are defined as:
\[
\theta_{\text{mrg}}(\beta) = \beta \cdot \theta_{\text{reg}} + (1 - \beta) \cdot \theta_{\text{glob}}
\]
where $\beta$ controls the interpolation between the region-specific and the original global model. For beta interpolation, we first select a certain value for the global-regional parity factor $\alpha$ (e.g., $\alpha = 0.65$) to balance region-specific performance and general capabilities. The optimal $\beta$ value is selected based on evaluation metrics:
{\small
\[
\beta^* = \underset{\beta \in [0.0, 1.0]}{\text{argmax}} \left[ (1 - \alpha) \cdot Q^{\mathcal{R}^{\text{reg}}}_{\theta_{mrg}(\beta)}+ \alpha \cdot Q^{\mathcal{R}^{\text{glob}}}_{\theta_{mrg}(\beta)} \right]
\]
}
where $Q^{\mathcal{R}^{\text{reg}}}_{\theta_{\text{mrg}}(\beta)}$ measures regional-specific performance from the evaluation sets $E^{\mathcal{R}^{\text{reg}}}$ and $Q^{\mathcal{R}^{\text{glob}}}_{\theta_{\text{mrg}}(\beta)}$ measures global performance from the evaluation sets $E^{\mathcal{R}^{\text{glob}}}$. This approach ensures minimal degradation of base model capabilities while achieving strong regional adaptation.

\begin{table*}[!t]
\centering
\resizebox{0.7\linewidth}{!}{
    \begin{tabular}{lccccccccc}
    \hline
    \multicolumn{1}{c}{\multirow{2}{*}{\textbf{Model}}} & \multicolumn{1}{c}{\multirow{2}{*}{\textbf{GRP}}}& \multicolumn{3}{c}{\textbf{Global}} & \multicolumn{4}{c}{\textbf{SEA-Specific}} \\
    \cmidrule(lr){3-5} \cmidrule(lr){6-9}
    &  & \textbf{Avg.} & \textbf{WC} & \textbf{CVQA} & \textbf{Avg.} & \textbf{SEAVQA} & \textbf{WC} & \textbf{CVQA} \\
    \hline
    \textbf{Google Gemma-3} & 59.4 & 63.5 & 59.8 & 67.2 & 56.3 & 41.0 & 60.1 & 67.8 \\
    \midrule
    \textbf{SEA-Gemma-3 5\%} & 64.0 & 64.3 & \textbf{60.0} & 68.7 & 63.7 & 61.2 & \textbf{60.3} & \textbf{69.5} \\
    \textbf{SEA-Gemma-3 10\%} & \textbf{64.1} & \textbf{64.4} & \textbf{60.0} & \textbf{68.8} & \textbf{63.8} & \textbf{61.7} & 60.2 & \textbf{69.5} \\
    \textbf{SEA-Gemma-3 50\%} & 57.3 & 56.7 & 51.6 & 61.8 & 57.8 & 59.5 & 51.4 & 62.6 \\
    \textbf{SEA-Gemma-3 70\%} & 56.1 & 56.3 & 51.9 & 60.6 & 56.0 & 54.0 & 52.6 & 61.3 \\
    \midrule
    \textbf{SEA-Gemma-3 (w/o merge)} & 42.2 & 42.1 & 48.5 & 35.6 & 42.2 & 41.9 & 48.6 & 36.2 \\
    \hline
    \end{tabular}
}
\caption{SEA-VLM evaluation results on CVQA, SEA-VQA, and WorldCuisine (WC). \textsc{<MODEL> (X\%)} denotes a merged model where X\% is the weight  $\beta$ use in the linear merging process of the region-specific model.}
\label{tab:sea-vlm-auto-eval}
\end{table*}

\begin{table*}[!t]
\centering
\resizebox{0.7\linewidth}{!}{
    \begin{tabular}{lcccccccc}
    \toprule
    \multicolumn{1}{c}{\multirow{2}{*}{\textbf{Model}}} & \multicolumn{3}{c}{\textbf{Overall}} & \multicolumn{5}{c}{\textbf{Language Breakdown}} \\
    \cmidrule(lr){2-4} \cmidrule(lr){5-9}
    & \textbf{GRP} & \textbf{Global} & \textbf{SEA} & \textbf{fil} & \textbf{ind} & \textbf{tha} & \textbf{vie} & \textbf{zsm} \\
    \hline
    \textbf{Google Gemma-3} & 2.29 & \textbf{2.54} & 2.09 & 1.69 & 2.15 & 2.17 & 2.37 & 2.07 \\
    \textbf{SEA-Gemma-3 10\%} & \textbf{2.31} & 2.42 & 2.22 & 1.88 & 2.07 & 2.29 & \textbf{2.61} & \textbf{2.25} \\
    \textbf{SEA-Gemma-3 (w/o merge)} & 1.74 & 1.18 & \textbf{2.23} & \textbf{2.75} & \textbf{2.29} & \textbf{2.33} & 1.76 & 2.00 \\
    \bottomrule
    \end{tabular}
}
\caption{SEA-VLM human evaluation results. We report the average rank across different models (Higher is better). \textsc{<MODEL> (X\%)} denotes a merge with X\% as the weight $\beta$ of the region-specific model.}
\label{tab:sea-vlm-human-eval}
\end{table*}

\section{Case Study on Southeast Asian (SEA) Adaptation}
\label{sec:experiment}

We explore GG-EZ on 3 distinct multimodal architectures: a large-scale Vision-Language Model (VLM) with 27B parameters, a contrastive Vision-Language Embedding Model (VL-Embed) with 1B parameters, and a Contextualized Diffusion Model (SEA-ImageGen) with 3.5B parameters. Our experiments focus on the SEA region, encompassing 11 countries -- Singapore, Indonesia, Malaysia, Brunei, Thailand, Philippines, Vietnam, Myanmar, Cambodia, Laos, and East Timor -- with a total population of approximately 700 million.  In this section, we provide detailed configurations for applying GG-EZ and the evaluation strategy used to assess its effectiveness.

\subsection{Data Curation}
\label{sec:data-curation_5_1}
For data curation, we leveraged prefiltered SEA-specific content from SEA-VL, culturally relevant imagery from CulturalGround \cite{nyandwi-etal-2025-grounding}, and translated instruction data from MAmmoTH-VL \cite{guo2024mammothvlelicitingmultimodalreasoning} (5 shards converted to major SEA languages, including Indonesian, Malaysian, Thai, Vietnamese, Filipino, Khmer, Lao, Chinese, and Tamil). To ensure high-quality translation on the underrepresented languages like Khmer and Lao, we ablate different translation models for each language and select the one with optimal quality and cost.\footnote{See Appendix~\ref{sec:translation-quality} for more detail.} The data filtering process applied a regional filter to isolate SEA-specific examples, utilized the UnifiedReward~\cite{wang2026unifiedrewardmodelmultimodal} model for quality assessment with a threshold set at 3 or above\footnote{We pick the best reward model based on our ablations on a human-annotated SEA-specific human preference dataset. The comparison of different reward model quality on the SEA regional test set is shown on Appendix~\ref{sec:reward-model}.}, and combined regional and translated instruction data to create the final fine-tuning dataset. This dataset curation is crucial to enable adaptation of a general global model into a strong regional-specific model as explained in \S\ref{sec:data-curation}.

\subsection{Regional Supervised Fine-tuning}

All models underwent supervised fine-tuning using the AdamW optimizer~\cite{Loshchilovetal2017} with linear learning rate decay. The SEA-adapted VLM (SEA-VLM) was trained from the Gemma-3 model as the initial checkpoint with a batch size of 64, learning rate of 2e-5, weight decay of 0.01, and 3 epochs. Before the supervised fine-tuning, we perform continuous pre-training on the Gemma-3 checkpoint using SEA-VL, XM3600, and Flickr30k, where we translated these datasets into SEA languages similar to Section~\ref{sec:data-curation_5_1}.  For the contextualized diffusion model (SEA-ImageGen), the model was fine-tuned from SDXL with a batch size of 32, learning rate of 1e-5, weight decay of 0.01, and 4 epochs. We only fine-tuned the U-Net module the model while keeping the VAE module as is. For the VL embedding model (SEA-VL Embed), the model was fully fine-tuned from SigLIP2-So400m with a batch size of 128, learning rate of 5e-6, 
and 2 epochs.  The resulting regionally-adapted models are denoted as SEA-Gemma-3, SEA-SDXL, and SEA-SigLIP2 for the VLM, diffusion, and embedding architectures respectively.

\begin{table*}[!t]
    \centering
    \resizebox{0.7\linewidth}{!}{
        \begin{tabular}{@{}lcccccc@{}}
        \toprule
        \multicolumn{1}{c}{\multirow{2}{*}{\textbf{Model}}} & \textbf{Overall} & \multicolumn{5}{c}{\textbf{Aspects}} \\
        \cmidrule(lr){3-7}
         & \textbf{Score} & \textbf{Attribute} & \textbf{Relation} & \textbf{Entity} & \textbf{Other} & \textbf{Global} \\
        \midrule
        \textbf{StabilityAI SDXL} & 73.75 & 79.18 & 86.38 & 81.07 & 60.00 & 83.89 \\
        \midrule
        \textbf{SEA-SDXL 25\%} & \textbf{74.75} & 80.43 & 86.34 & \textbf{81.86} & 61.60 & 85.41 \\
        \textbf{SEA-SDXL 50\%} & 74.61 & \textbf{80.93} & \textbf{87.12} & 81.15 & 60.00 & \textbf{86.02} \\
        \textbf{SEA-SDXL 75\%} & 74.39 & 79.98 & 85.96 & 81.36 & 63.20 & 82.67 \\
        \midrule
        \textbf{SEA-SDXL (w/o merge)} & 74.32 & 80.79 & 86.92 & 81.57 & \textbf{65.20} & 81.16 \\
        \bottomrule
        \end{tabular}
    }
    \caption{SEA-SDXL Performance on Image Generation Benchmark DPGBench (Higher is better). \textsc{<MODEL> (X\%)} denotes a linear merge where X\% is the weight  $\beta$ of the region-specific model.}
    \label{tab:dpgbench_scores}
\end{table*}

\subsection{Global-Regional Merging}

After fine-tuning, we performed linear model merging to combine region-specific adaptations with the original base models. For VL-Embed and SEA-ImageGen, we explored interpolation weights $\beta$ in the range [0.25, 0.5, 0.75], while for the SEA-VLM we tested a broader range of $\beta$ values [0.05, 0.10, 0.5, 0.7]. The optimal $\beta$ was determined by maximizing a weighted combination of regional and general performance metrics, where RegionAcc measures performance on SEA-specific validation data and GeneralAcc evaluates performance on broad-domain benchmarks. This formulation prioritizes regional adaptation while ensuring minimal degradation of foundational capabilities. We denote the resulting merged model as ``\textsc{<MODEL> (X\%)}'' -- e.g., SEA-Gemma-3 10\% --  where \textsc{X\%} denotes the weight percentage $\beta$ of the regionally-specific model used in the linear merging.

\subsection{Evaluation}

For the SEA-VLM model, evaluation included the SEAVQA regional visual question answering benchmark, CVQA for culturally nuanced queries, WorldCuisine for food identification tasks, and human evaluation to assess cultural relevance and accuracy.\footnote{Due to limited evaluation resources for the region, we develop two human evaluation sets: (1) SEA-AYA a human translated AyaVisionBench~\cite{dash2026aya}; and (2) SEA-VL VQA, a human annotated VQA prompts from SEA imagery. Details in Appendix~\ref{sec:human-test-set}.} The SEA-VL Embed was evaluated on SEAVQA and CVQA to measure regional and cultural performance in embedding space. The SEA-ImageGen model underwent assessment using a  DPGBench benchmark~\cite{hu2024ellaequipdiffusionmodels} as a proxy for the global quality of the models and diffusion model human evaluation set from SEA-VL~\cite{cahyawijaya-etal-2025-crowdsource} to gauge cultural relevance and visual quality on the SEA region. All evaluations were conducted on held-out test sets to ensure fair comparison across architectures and adaptation strategies. For human evaluation we use a 3-level likert scoring with native speaker annotators for each of the evaluated regional language. Finally, to compute a fair and practical global-regional parity (GRP) score from both of the global and region-specific evaluations, we set the globalization factor $\alpha = 0.43$\footnote{$\alpha = 0.43$ indicates a moderate tendency toward SEA-focused quality $Q^{\mathcal{R}^{\text{reg}}}$ with a fair amount of consideration for global quality $Q^{\mathcal{R}^{\text{glob}}}$.} derived from 2023 KOF Globalization Index~\cite{Haelg2019,Gygli2019} which measures the level of globalization across 190 countries.

\begin{table}[!t]
\centering
\resizebox{\linewidth}{!}{
    \begin{tabular}{lccccccccc}
    \toprule
    \multicolumn{1}{c}{\multirow{2}{*}{\textbf{Model}}} & \multicolumn{4}{c}{\textbf{Correctness}} & \multicolumn{4}{c}{\textbf{Naturalness}} \\
    \cmidrule(lr){2-5} \cmidrule(lr){6-9}
    & \textbf{Overall} & \textbf{T} & \textbf{L} & \textbf{C} & \textbf{Overall} & \textbf{T} & \textbf{L} & \textbf{C} \\
    \midrule
    \textbf{StabilityAI SDXL} & 1.491 & 1.470 & 1.636 & 1.387 & 1.675 & 1.436 & 2.023 & 1.613 \\
    \textbf{SEA-SDXL 25\%} & \textbf{1.569} & \textbf{1.587} & \textbf{1.729} & \textbf{1.413} & \textbf{1.767} & \textbf{1.473} & \textbf{2.124} & \textbf{1.753} \\
    \textbf{SEA-SDXL  (w/o merge)} & 1.431 & 1.473 & 1.527 & 1.307 & 1.557 & 1.340 & 1.806 & 1.560 \\
    \bottomrule
    \end{tabular}
}
\caption{SEA-SDXL human evaluation results on 3 distinct cultural aspects. \textbf{``T''} denotes Tradition, \textbf{``L''} denotes Landmark,  and \textbf{``C''} denotes Cuisine}
\label{tab:imagegen_human_metrics}
\end{table}

\section{Result and Discussion}
\label{sec:result}

\subsection{Overall Results}

\begin{table*}[!t]
\centering
\resizebox{0.7\linewidth}{!}{
    \begin{tabular}{lcccccccc}
    \toprule
    \multicolumn{1}{c}{\multirow{2}{*}{\textbf{Model}}} & \multicolumn{1}{c}{\multirow{2}{*}{\textbf{GRP}}} & \multicolumn{3}{c}{\textbf{CVQA}} & \multicolumn{4}{c}{\textbf{SEAVQA}} \\
    \cmidrule(lr){3-5} \cmidrule(lr){6-9}
     & & \textbf{Global} & \textbf{SEA} & \textbf{Others} & \textbf{SEA} & \textbf{Khm} & \textbf{Ind} & \textbf{Vie} \\
    \midrule
    \textbf{Google SigLIP2} & 25.17 & 25.51 & 24.02 & 25.84 & 25.81 & 28.95 & 25.00 & 25.56 \\
    \midrule
    \textbf{SEA-SigLIP2 25\%} & 24.96 & 24.38 & 24.44 & 24.36 & 26.36 & 27.63 & 25.40 & 23.96 \\
    \textbf{SEA-SigLIP2 50\%} & 27.10 & \textbf{27.52} & 25.50 & \textbf{27.97} & 28.06 & \textbf{31.25} & 26.99 & 27.16 \\
    \textbf{SEA-SigLIP2 75\%} & \textbf{27.96} & 27.12 & \textbf{27.51} & 27.03 & \textbf{29.66} & 28.62 & \textbf{30.05} & 28.75 \\
    \midrule
    \textbf{SEA-SigLIP2 (w/o merge)} & 26.96 & 26.75 & 25.29 & 27.07 & 28.96 & 28.29 & 28.59 & \textbf{29.07} \\
    \bottomrule
    \end{tabular}
}
\caption{SEA-VL Embedding Zero-Shot Performance on Cultural VQA Benchmarks. \textsc{<MODEL> (X\%)} denotes a linear merge where X\% is the weight $\beta$ of the region-specific model. For SEA-VQA, we provide additional breakdown of 3 of the biggest country data in the dataset with ``Khm''=Cambodia, ``Ind''=Indonesia, ``Vie''=Vietnam.}
\label{tab:sea_vl_embed_eval}
\end{table*}

\subsubsection{GG-EZ on SEA-VLM}

We observe two major trends in SEA-Gemma-3 performance (Table~\ref{tab:sea-vlm-auto-eval}). First, without model merging, SEA-Gemma-3 generally underperforms Gemma-3 on CVQA and World-Cuisine benchmarks, likely due to reduced task generalization after regional fine-tuning. However, it exceeds Gemma-3 on the SEA-specific SEA-VQA evaluation. This suggests that pure supervised fine-tuning for regional adaptation may not be optimal, as pre-trained VLMs already contain diverse data that can be compromised by smaller regional datasets. With model merging, performance improves significantly across all evaluations. A 50:50 mix (SEA-Gemma-3 50\%) achieves a turning point where global performance slightly decreases while SEA-specific performance surpasses Gemma-3. Increasing the merging weight to 10\% (SEA-Gemma-3 10\%) further improves results, outperforming Gemma-3 on all evaluation sets with a 1\% global performance gain and 7.5\% SEA-specific improvement, yielding the highest GRP. This demonstrates GG-EZ's ability to create regionally-adapted models strong in both specificity and generalization.

Human evaluation (Table~\ref{tab:sea-vlm-human-eval}) confirms these trends. SEA-Gemma-3 (w/o merge) shows superior regional specialization with the best average ranks across Southeast Asian languages: Filipino (2.75), Indonesian (2.29), Thai (2.33), and SEA region (2.23). Gemma-3 maintains global dominance (2.54), while SEA-Gemma-3 10\% achieves the best Vietnamese (2.61) and Malaysian (2.25) scores with minimal global degradation (2.42 vs. 2.54), yielding the highest GRP of 2.31. This confirms that even modest SEA integration provides measurable regional benefits (Appendix~\ref{sec:detailed-results} and ~\ref{sec:sample-output}).

\subsubsection{GG-EZ on SEA-ImageGen}

The SEA-SDXL models demonstrate strong regional adaptation while maintaining robust global performance on DPGBench (Table~\ref{tab:dpgbench_scores}). The fully fine-tuned SEA-SDXL achieves 74.32 overall, slightly improving upon SDXL's 73.75. GG-EZ's linear merging further enhances performance, with SEA-SDXL 25\% and 50\% models reaching 74.75 and 74.61 respectively, surpassing both the general model and regional-specific SEA-SDXL.

Human evaluation results (Table~\ref{tab:imagegen_human_metrics}) confirm the merging strategy's effectiveness for regional context. The SEA-SDXL 25\% model outperforms both SDXL and fully fine-tuned SEA-SDXL across all cultural aspects (Tradition, Landmark, Cuisine) in both correctness and naturalness metrics. This indicates the merging process successfully recovers generalization capability lost during regional specialization, enabling the models to maintain or enhance global performance while excelling in regional context evaluations.\footnote{GRP score was not measured for image generation due to lack of comparable test sets, though the model clearly improves both regional-specific and general image generation scores.}

\begin{figure}[!t]
    \centering
    \includegraphics[width=\linewidth, trim=0 0.4cm 0 0, clip]{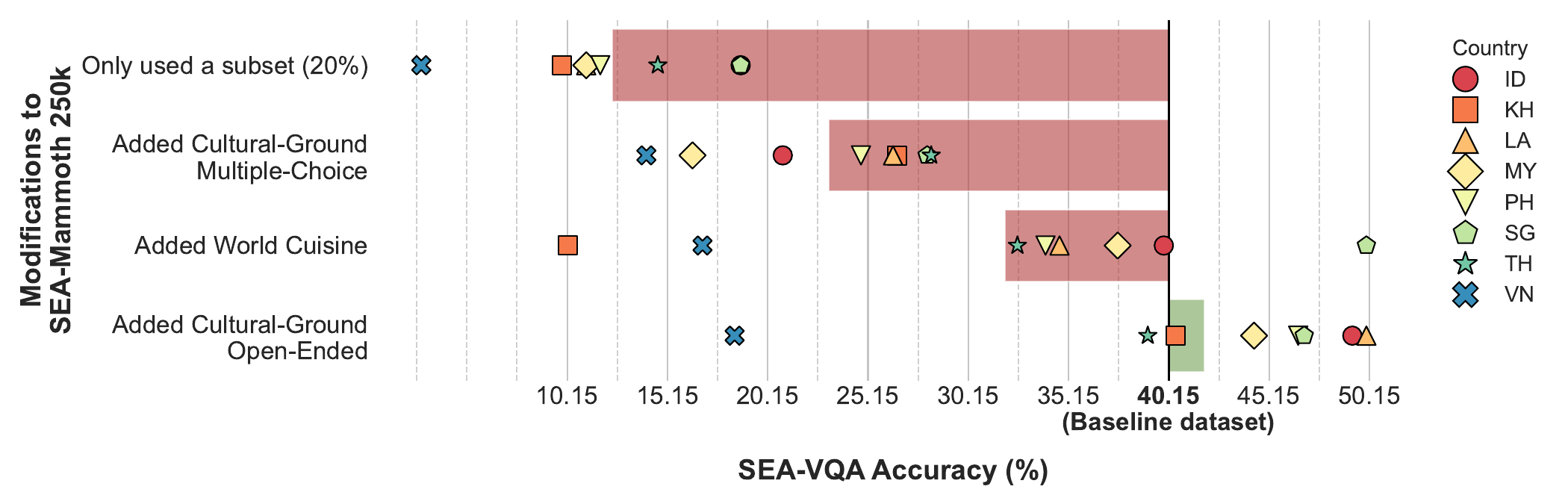}
    \caption{Impact of regional-specific data curation strategy on SEA-Gemma-3.
    }
    \label{fig:data_ablation}
\end{figure}

\begin{figure*}[!t]
    \centering
    \begin{subfigure}{0.39\textwidth}
        \centering
        \includegraphics[width=1.0\linewidth]{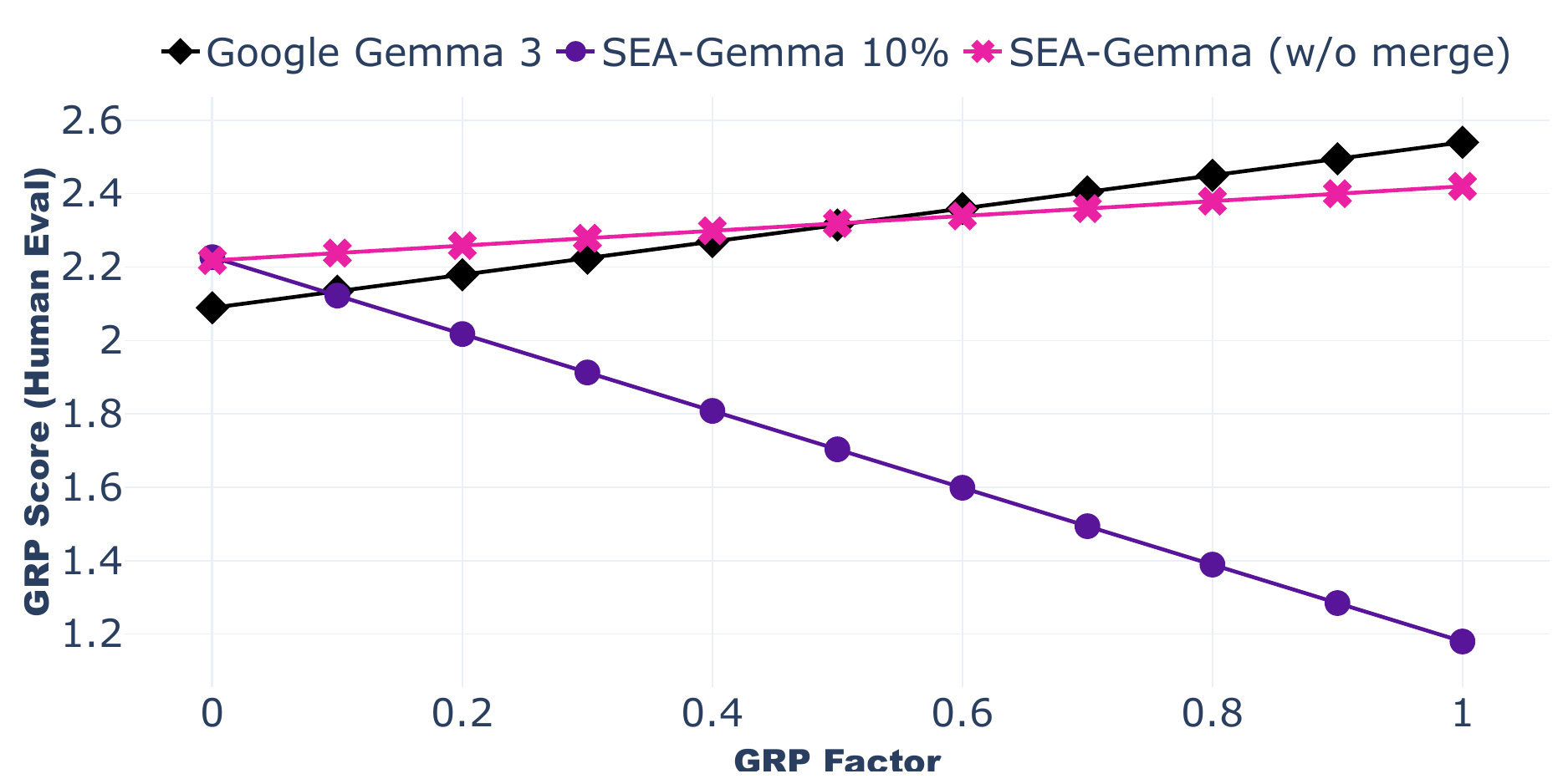} 
        \includegraphics[width=1.0\linewidth]{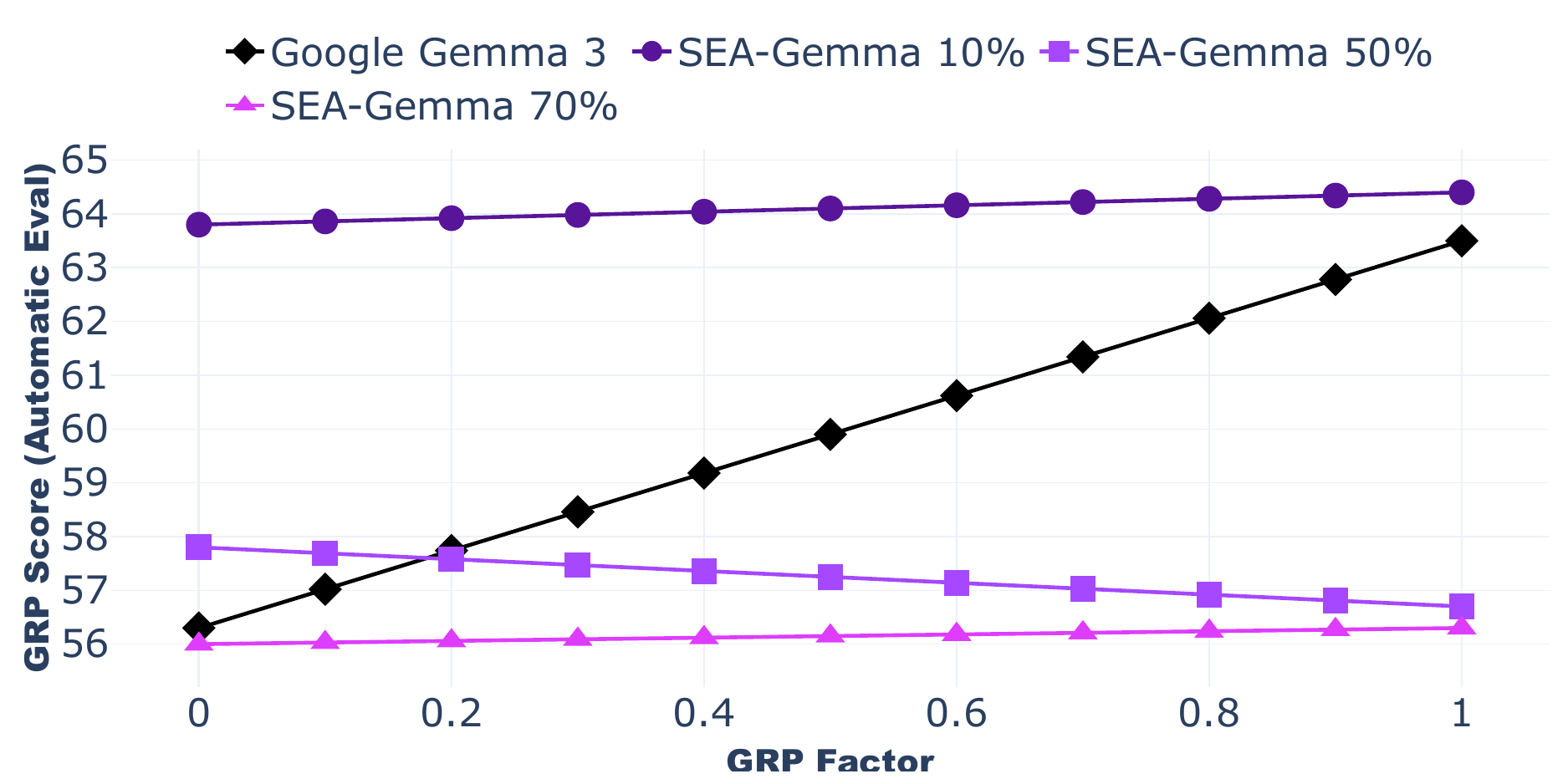}
    \end{subfigure}
    \begin{subfigure}{0.58\textwidth}
        \centering
        \includegraphics[width=1.0\linewidth]{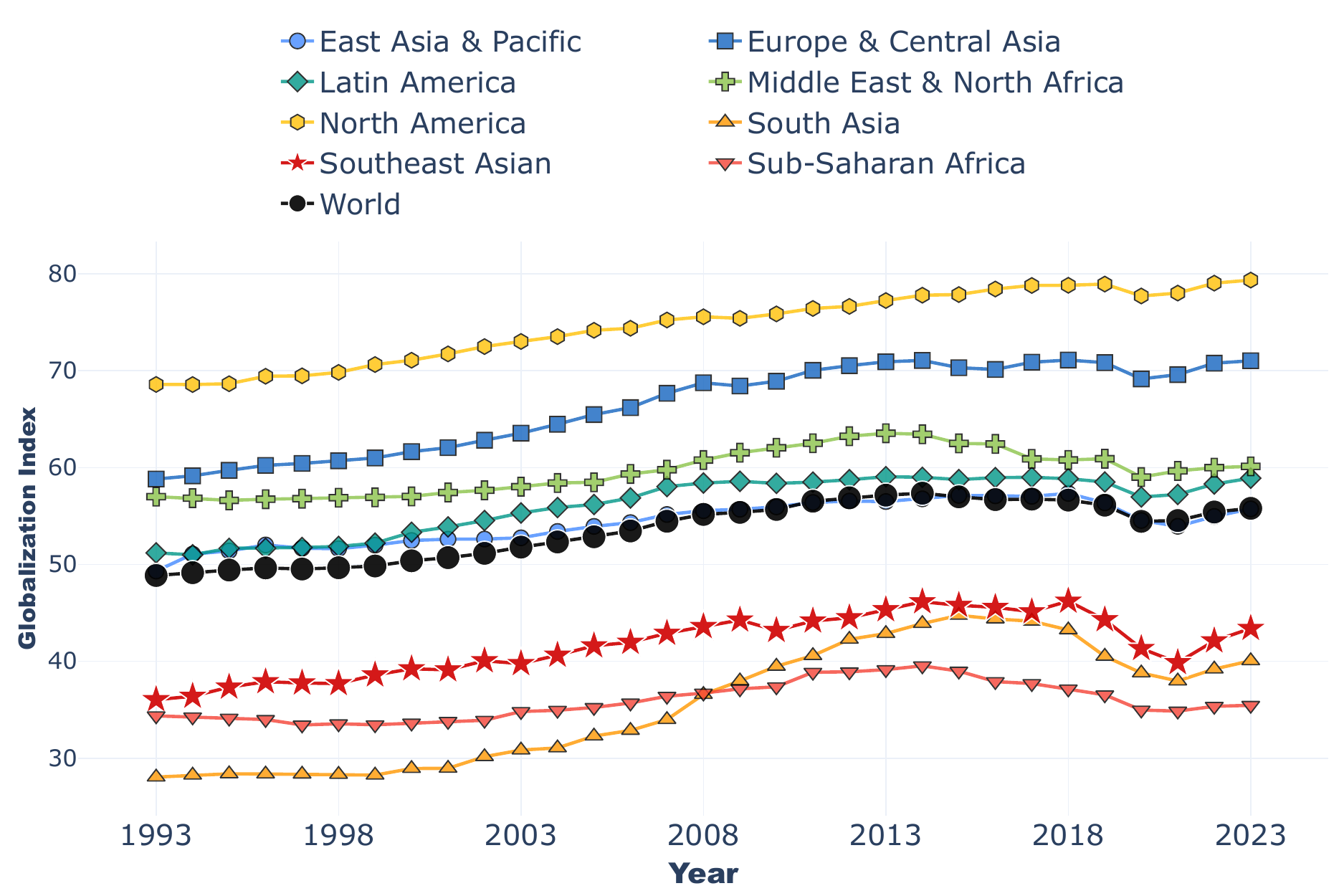}
    \end{subfigure}
    \caption{\textbf{(left)} Impact of globalization factor $alpha$ to GRP across different models. Optimizing on a misaligned $\alpha$ can lead to suboptimal performance. \textbf{(right)} We derive $\alpha$ from the KOF globalization index~\cite{Haelg2019,Gygli2019} to better reflect the degree of globalization across regions. The globalization index is distinct across regions and evolves over time.}
    \label{fig:kof-index}
\end{figure*}

\subsubsection{GG-EZ on SEA-VL Embed}

The SEA-SigLIP2 model outperforms the original Google SigLIP2 on both SEA and non-SEA regions in CVQA and SEA-VQA evaluations (Table~\ref{tab:sea_vl_embed_eval}). Merging SEA-SigLIP2 with the original SigLIP2 further enhances performance. The SEA-SigLIP2 50\% model achieves the highest global CVQA score (27.52) and strongest non-SEA performance (27.97), excelling in general knowledge transfer. The SEA-SigLIP2 75\% model demonstrates superior regional specialization with the highest SEA-VQA overall score (29.66), peak performance for Indonesia (30.05) and Vietnam (28.75), and maintains strong global performance (27.12), yielding the highest GRP score (27.96) compared to the original SigLIP2 (25.17) and fine-tuned SEA-SigLIP2 (26.96). The merging approach effectively balances regional adaptation and generalization, preserving and enhancing SigLIP2's broad capabilities while incorporating specialized SEA knowledge.\footnote{Detailed evaluation breakdowns are in Appendix~\ref{sec:detailed-results}.}

\subsection{Impact of Regional Quality Filtering}
\label{sec:data-curation}

Data quality and relevance critically impact region-specific model adaptation performance~\cite{cahyawijaya-etal-2023-nusawrites,cahyawijaya-etal-2024-cendol}. SEA-Gemma-3 ablation studies show naive augmentation does not guarantee gains; outcomes depend on data structure, scope, and provenance. As in Figure~\ref{fig:data_ablation}, data volume is a bottleneck: a 20\% subset of the SEA-Mammoth 250k dataset (translated to 10 languages) causes a 70\% performance drop vs. full data, stressing the need for sufficient scale. 

Integrating the SEA-filtered Cultural-Ground dataset~\cite{nyandwi-etal-2025-grounding} shows performance depends on VQA format. Open-ended VQA boosts regional understanding to 41.9\% (baseline) but is sensitive to prompts; multiple-choice degrades to 21.6\%. The SEA-filtered WorldCuisine dataset~\cite{winata-etal-2025-worldcuisines} reduces performance by ~42\%. We hypothesize its regional cuisine focus narrows the model, lacking comprehensive cultural knowledge (e.g., landmarks, history, clothing). Thus, strategic curation is vital; indiscriminate dataset addition without format- and domain-aware evaluation harms generalizability.

\subsection{Globalization Factor in Anthropogenic Regional Adaptation}
\label{sec:global-regional-parity}

The globalization factor $\alpha$ governs GRP optimization for Anthropogenic Regional Adaptation, determining how models should balance global and regional characteristics. An appropriate $\alpha$ reflects actual regional interaction patterns, whether societies prefer unified global perspectives or localized regional views. Misalignment causes undesirable model behavior (Figure~\ref{fig:kof-index}, left). We derive $\alpha$ from the KOF-GI (Konjunkturforschungsstelle Globalization Index) \cite{Haelg2019,Gygli2019}, using the de facto interpersonal component that measures person-to-person globalization intensity \cite{martens2009maastricht,axel2010measuring,figge02112014}. This provides a standardized, quantifiable metric by calculating KOF-GI expectations across target region countries $\mathcal{R}^{\text{reg}}$.

The KOF-GI-based $\alpha$ offers two key advantages: 1) dynamic adaptation as globalization evolves over time (Figure~\ref{fig:kof-index}, right), and 2) an objective, external metric reducing subjectivity in regional importance determination. This establishes a fair global-regional parity baseline that bridges the gap between purely global models (lacking regional specificity) and purely regional models (lacking global coherence) (Figure~\ref{fig:highlight-figure}). Models optimized with KOF-GI-derived $\alpha$ adaptively balance focus: emphasizing regional performance in less globalized areas while maintaining global consistency in highly interconnected regions. This creates a system that simultaneously satisfies universal applicability and specialized excellence demands, enabling robust performance across heterogeneous regional contexts.\footnote{For detailed KOF-GI breakdown, see Appendix~\ref{sec:kof-gi}.}

\section{Conclusion}

We propose the Anthropogenic Regional Adaptation framework to address cultural insensitivity and reduced performance in underrepresented regions, establishing a foundational paradigm for human-centric alignment in vision-language systems. Our GG-EZ approach implements this framework, demonstrating 5-15\% gains in cultural relevance metrics across SEA contexts (Table 1) while maintaining or improving generalization capabilities across three architectures -- Gemma-3 27B, SDXL, and SigLIP-2 -- enabling equitable deployment without sacrificing core performance. This work highlights GRP's importance in human-centric alignment, providing data-driven and dynamically adjusted optimization.

\section{Limitation}

Our study demonstrates promising gains in regional cultural relevance, but several constraints limit the breadth and depth of the findings. First, the experiments are confined to relatively modest model sizes (Gemma‑3 27B, SigLIP‑2, and SDXL), which may not capture the full spectrum of performance improvements achievable with larger architectures; scaling to billion‑plus parameter models could reveal different trade‑offs between regional adaptation and global generalization. Second, while we target eleven Southeast Asian languages, the coverage is uneven: major languages such as Indonesian, Thai, and Vietnamese receive substantially more data and evaluation than smaller languages like Khmer and Lao, potentially biasing the results and leaving gaps in truly low‑resource linguistic contexts. Third, the exploration of model merging strategies is limited to linear interpolation with a few discrete weight points; more sophisticated merging techniques remain underexplored, and could yield better parity or higher regional gains without sacrificing global performance.

\section{Ethical Statement}

All human annotations were conducted on a voluntary basis by the authors and members of relevant regional communities, with fair compensation provided for their contributions. Annotations underwent rigorous double-checking to ensure quality, safety, and cultural relevance, adhering to ethical guidelines for data collection in underrepresented regions. The evaluation design incorporated a fair scoring system to minimize bias, and annotators were selected from diverse linguistic and cultural backgrounds within Southeast Asia to ensure contextual accuracy. No sensitive or private data was used, and all processes aligned with principles of informed consent and equitable participation. During writing of the paper, we also use AI assistant to help with paraphrasing and post-editing.

\section{Acknowledgements}

We extend our gratitude to all affiliated institutions, particularly Oracle and AISingapore, for providing the compute resources essential to this project. We express our deepest appreciation to Sarana Nutanong, Genta Indra Winata, Fajri Koto, Derry Wijaya, Muhamad Risqi U Saputra, Gholamreza Haffari, Enshiun Annie Lee, and William Tjhi for their invaluable assistance throughout the various phases of this project, including idea discussion, information dissemination, financial support, and other invaluable supports. Additionally, we wish to acknowledge Daryl Peralta, Robert Wijaya, Pume Tuchinda, and Ong Lip Wei for their dedicated support and expertise in dataset annotations and human evaluations. We acknowledge the support of the Google Cloud Research Credits program (Award No. GCP19980904), which provided cloud computing credits used for data set translation. This research is also supported by the National Research Foundation, Singapore, under its National Large Language Models Funding Initiative. Any opinions, findings, and conclusions or recommendations expressed in this material are those of the author(s) and do not reflect the views of the National Research Foundation, Singapore.

%
%
\bibliography{custom}

\clearpage

\appendix

\renewcommand{\thesection}{\Alph{section}} 
\setcounter{section}{0} 

\section{Assessment for SEA languages Translation Quality}
\label{sec:translation-quality}

\begin{table}[!h]
\centering
\resizebox{1.0\linewidth}{!}{
    \begin{tabular}{llcc}
    \hline
    \textbf{Language} & \textbf{Model} 
    & \textbf{Grammar} 
    & \textbf{Naturalness} \\
    \hline
    \multirow{4}{*}{Burmese (mya)} 
    & NLLB &   4.6 $\pm$ 0.8 & 4.1 $\pm$ 1.1 \\
    & Gemma-3-27b &   4.1 $\pm$ 1.2 & 3.6 $\pm$ 1.2 \\
    & Gemini-2.5-flash &  4.8 $\pm$ 0.6 & 4.6 $\pm$ 0.8 \\
    & Gemini-2.5-pro & \textbf{4.9 $\pm$ 0.4} & \textbf{4.9 $\pm$ 0.5} \\
    \hline
    \multirow{4}{*}{Thai (tha)} 
    & NLLB &   3.3 $\pm$ 1.1  & 3.0 $\pm$ 1.2 \\
    & Gemma-3-27b &   4.2 $\pm$ 0.9 & 4.1 $\pm$ 0.9 \\
    & Gemini-2.5-flash & \textbf{4.8 $\pm$ 0.5} & \textbf{4.7 $\pm$ 0.5}  \\
    & Gemini-2.5-pro  & 4.7 $\pm$ 0.5 & 4.6 $\pm$ 0.5 \\
    \hline
    \multirow{4}{*}{Filipino (fil)}
    & NLLB &  4.4 $\pm$ 0.6  & 4.3 $\pm$ 0.9  \\
    & Gemma-3-27b & \textbf{4.8 $\pm$ 0.6}  & \textbf{4.6 $\pm$ 0.8}\\
    & Gemini-2.5-flash & 4.7 $\pm$ 0.4  & 4.4 $\pm$ 0.7 \\
    & Gemini-2.5-pro  & \textbf{4.8 $\pm$ 0.4} & 4.5 $\pm$ 0.6 \\
    \hline
    \multirow{4}{*}{Indonesian (ind)} 
    & NLLB & 4.3 $\pm$ 0.9  & 3.8 $\pm$ 1.1 \\
    & Gemma-3-27b &  \textbf{4.6 $\pm$ 0.6}  & 4.2 $\pm$ 0.9 \\
    & Gemini-2.5-flash & 4.5 $\pm$ 0.6  & 4.1 $\pm$ 0.9 \\
    & Gemini-2.5-pro & \textbf{4.6 $\pm$ 0.5}  &\textbf{4.3 $\pm$ 0.8}  \\
    \hline
    \multirow{4}{*}{Vietnamese (vie)}
    & NLLB &   4.7 $\pm$ 0.5 & 4.4 $\pm$ 0.8 \\
    & Gemma-3-27b &  \textbf{5 $\pm$ 0.1} & 4.9 $\pm$ 0.3 \\
    & Gemini-2.5-flash &  \textbf{5 $\pm$ 0.1}  & 4.9 $\pm$ 0.2 \\
    & Gemini-2.5-pro &  \textbf{5 $\pm$ 0.1}  & \textbf{5 $\pm$ 0.1} \\
    \hline
    \end{tabular}
}
\caption{Human evaluation of English→5 SEA languages translation quality using a five-point scale (mean ± standard deviation) }
\label{tab:systematic_eval}
\end{table}

We conducted a systematic evaluation on 100 randomly selected human-generated English captions from the SEA-VL \cite{cahyawijaya-etal-2025-crowdsource} dataset for Burmese (mya), Thai (tha), Filipino (fil), Indonesian (ind), and Vietnamese (vie). The evaluation employed a five-point scale (1–5) assessing grammatical accuracy and naturalness, using NLLB \cite{nllb2022}, Gemma-3-27b \cite{gemma3_2025}, Gemini-2.5-flash, and Gemini-2.5-pro \cite{gemini2023} models. Table~\ref{tab:systematic_eval} presents the results of the translation quality evaluation in five Southeast Asian languages using a five-point scale (mean ± standard deviation). Overall, the Gemini-2.5 models consistently achieve the highest scores for Burmese (mya) and Thai (tha), with Gemini-2.5-Pro obtaining the best performance for Burmese and Gemini-2.5-Flash performing slightly better for Thai. For Filipino (fil), Indonesian (ind), and Vietnamese (vie), Gemma-3-27B achieves competitive or top performance, often matching or exceeding the Gemini models in grammatical accuracy while maintaining high naturalness scores.
Based on these findings, Gemma-3-27b was used for Indonesian (ind), Vietnamese (vie), Standard Malay (zsm), Filipino (fil), and Chinese (zho), where it demonstrated strong and stable performance. Gemini-2.5-flash was used for Thai (tha), Burmese (mya), Lao (lao), Khmer (khm), and Tamil (tam), as the Gemini family showed superior robustness and higher translation quality for these languages.

\section{Reward Model Ablation}
\label{sec:reward-model}
We evaluated four reward models: ImageReward~\cite{xu2023imagereward}, HPSv2~\cite{wu2023humanpreferencescorev2}, VisionReward-Image~\cite{xu2026visionrewardfinegrainedmultidimensionalhuman}, and UnifiedReward~\cite{wang2026unifiedrewardmodelmultimodal}. In order to pick a reward model to use for data curation as described in section~\ref{sec:data-curation_5_1}, we measured how much the reward models agreed with human evaluations from SEA-VL~\cite{cahyawijaya-etal-2025-crowdsource}. For each image $I_n$, we normalized the scores via min-max normalization and then took the average across categories to get a human annotator score $s^n_{human} \in [0,1]$. Notably, the absolute score range of ImageReward-style models varies across datasets, making direct normalization and comparison with human ground-truth scores unreliable. Therefore, instead of comparing absolute scores, we reformulated the evaluation as a pairwise preference task. Specifically, we randomly sampled 500 pairs of images $I_a$ and $I_b$ such that $s^a_{human} > s^b_{human}$. For each reward model, we then compute the rate at which it gives $I_a$ a higher score than $I_b$ i.e. $s^a_{rm} > s^b_{rm}$. Results are summarized in table~\ref{tab:reward_model_evals}. Although VisionReward-Image achieved the highest pairwise agreement with human judgments, we ultimately selected UnifiedReward by considering both computational efficiency and predictive performance. While its accuracy is slightly lower, UnifiedReward offers a more favorable trade-off overall, as it supports efficient batched inference through vLLM~\cite{kwon2023efficient}, substantially improving throughput for large-scale data filtering.

\begin{table}[h!]
\centering
\resizebox{\linewidth}{!}{
    \begin{tabular}{lc}
    \hline
    \textbf{Reward Model} & \textbf{Rate where $s^a_{rm} > s^b_{rm}$} \\
    \hline
    ImageReward & 0.394 \\
    HPSv2 & 0.384 \\
    VisionReward-Image & \textbf{0.466} \\
    UnifiedReward & 0.442 \\
    \hline
    \end{tabular}
}
\caption{Rate at which $s^a_{rm} > s^b_{rm}$ for each reward model under consideration.}
\label{tab:reward_model_evals}
\end{table}

\section{Annotated Human Evaluation Test Sets}
\label{sec:human-test-set}

We conduct human evaluation on two datasets: SEA-VL VQA and SEA AYA.

\subsection{SEA-VL VQA}

SEA-VL VQA contains $\sim$1.1k visual questions spanning 9 official languages in Southeast Asia: Malay (\texttt{zsm}), Vietnamese (\texttt{vie}), Thai (\texttt{tha}), Indonesian (\texttt{ind}), Filipino (\texttt{fil}), Tamil (\texttt{tam}), Khmer (\texttt{khm}), Chinese (\texttt{cmn}), and Burmese (\texttt{mya}). The culturally grounded images were originally collected via crowdsourcing by~\newcite{cahyawijaya-etal-2025-crowdsource}. For each image, a native speaker authored a visual question in the target language, which was subsequently reviewed by two additional native speakers, with revisions made when needed to ensure quality.

To promote consistency and cultural relevance, we provided annotators with detailed instructions for writing high-quality visual questions. Specifically, questions were required to: (1) be directly grounded in the visual content of the image; (2) go beyond low-level perception and instead target culturally salient elements, such as traditional attire, local cuisine, religious practices, architectural styles, or social activities; (3) be clear, grammatically correct, and contextually appropriate; and (4) be answerable solely from the image.

We also specified several categories of questions to avoid. These included: (1) overly generic questions (e.g., "What is this?" or "What color is the shirt?"); (2) purely factual questions lacking cultural grounding (e.g., "How many people are in the image?"); (3) questions that cannot be answered from the image alone (e.g., "What is her name?" unless explicitly shown); (4) questions that fail to reflect the Southeast Asian context without clear justification; and (5) questions that can be answered without the visual context.

\subsection{SEA AYA}

We construct SEA AYA by manually translating 135 visual questions from the Aya Vision Benchmark~\cite{ayavisionbench} into 6 official Southeast Asian languages: Thai (\texttt{tha}), Malay (\texttt{zsm}), Filipino (\texttt{fil}), Tamil (\texttt{tam}), Chinese (\texttt{cmn}), and Burmese (\texttt{mya}). Combined with the English (\texttt{eng}) and Indonesian (\texttt{ind}) subsets already available in Aya Vision Benchmark, the resulting benchmark contains $\sim$1.2k instances covering 9 vision-language tasks: image captioning, chart and figure understanding, image difference detection, general VQA, OCR, document understanding, text transcription, visual reasoning, and screenshot-to-code generation.
For each example, annotators were first given a rough machine-translated draft as a starting point. They were then asked to revise it to ensure that the final translation is fluent, culturally appropriate, and semantically faithful to the English source.

For quality and consistency, our annotation guidelines emphasized several principles: (1) preserve the meaning, tone, and intent of the original English prompt; (2) produce text that is natural and fluent for native speakers of the target language; (3) use grammar, vocabulary, and syntax that conform to standard usage in the target language; (4) match the level of formality of the source; (5) respect language-specific cultural and pragmatic norms, including idiomatic usage and politeness conventions; and (6) appropriately translate, retain, or normalize named entities and technical terms when necessary.

We also requested annotators to avoid common translation errors, including semantic drift, omission of essential information, overly literal adherence to machine-translated phrasing, literal translation of idiomatic expressions, over-translation or under-translation, awkward or unidiomatic wording, and inconsistent terminology, particularly across multi-sentence examples.

\section{Comparison of SEA-Gemma-3 and Other VLMs}
\label{sec:competitors-comparison}

We provide the full comparison per language on CVQA and per region on SEA-VQA in Table~\ref{tab:cvqa-competitor} and Table~\ref{tab:sea-vqa-competitor}, respectively.
The merged global-regional model SEA-Gemma-3 10\% achieves strongest performance on SEA regions and other regions on CVQA, producing a solid performance even higher compared to other smaller and similar sized models like Llama-3.2 Vision~\cite{grattafiori2024llama3herdmodels}, Pixtral~\cite{agrawal2024pixtral12b}, Gemma-3~\cite{gemma3_2025}, and even outperform larger models such as Aya-Vision (32B)~\cite{dash2026aya}, Qwen-3 VL (32B)~\cite{bai2025qwen3vltechnicalreport}, Qwen-2.5-VL (72B)~\cite{bai2025qwen25vltechnicalreport}, and Molmo (72B)~\cite{deitke2024molmopixmoopenweights}. Similarly on SEA-VQA, SEA-Gemma-3 10\% also achieves the best average regional score, although on several regions it is still outperformed by large models especially Aya-Vision (32B)~\cite{dash2026aya} which showcase strong performance on multiple SEA regions but falls short on Indonesia, Thailand, and Laos.

\begin{table*}[!t]
    \centering
    \resizebox{\linewidth}{!}{
        \begin{tabular}{lccccccccc}
        \toprule
         & \textbf{Llama-3.2} & \textbf{Pixtral} & \textbf{Gemma-3} & \textbf{SEA-Gemma-3} & \textbf{SEA-Gemma-3} & \textbf{Aya-Vision} & \textbf{Qwen-3 VL} & \textbf{Qwen-2.5-VL} & \textbf{Molmo} \\
        \textbf{Language} & \textbf{Vision (11B)} & \textbf{(12B)} & \textbf{(27B)} & \textbf{(27B)} & \textbf{10\% (27B)} & \textbf{(32B)} & \textbf{(32B)} & \textbf{(72B)} & \textbf{(72B)} \\
        \midrule
        \textbf{Indonesian} & 56.31 & 62.86 & 66.50 & 21.60 & 66.26 & 67.88 & 66.26 & 66.50 & 63.83 \\
        \textbf{Filipino} & 51.72 & 64.53 & 66.01 & 24.14 & 70.94 & 66.34 & 64.04 & 65.02 & 64.53 \\
        \textbf{Malay} & 56.19 & 61.90 & 67.62 & 29.84 & 71.43 & 72.38 & 70.16 & 62.50 & 69.84 \\
        \textbf{Chinese} & 62.80 & 68.93 & 72.66 & 40.73 & 75.14 & 77.48 & 78.28 & 80.98 & 79.29 \\
        \textbf{Tamil} & 58.41 & 51.87 & 73.36 & 40.65 & 74.30 & 61.68 & 61.68 & 58.88 & 58.41 \\
        \textbf{Javanese} & 47.81 & 51.18 & 62.96 & 22.22 & 63.64 & 55.56 & 54.55 & 55.22 & 54.88 \\
        \textbf{Minangkabau} & 51.79 & 51.39 & 64.54 & 23.90 & 67.73 & 61.75 & 54.98 & 51.79 & 58.17 \\
        \textbf{Sundanese} & 44.00 & 49.00 & 59.00 & 12.50 & 63.50 & 52.53 & 55.50 & 56.50 & 52.00 \\
        \midrule
        \textbf{SEA Avg.} & 53.63 & 57.71 & 66.58 & 26.95 & 69.12 & 64.45 & 63.18 & 62.17 & 62.62 \\
        \midrule
        \textbf{Amharic} & 39.32 & 32.91 & 62.82 & 26.07 & 62.39 & 29.18 & 42.74 & 36.48 & 45.30 \\
        \textbf{Bengali} & 55.59 & 48.25 & 72.03 & 33.57 & 75.87 & 64.31 & 64.34 & 61.27 & 68.88 \\
        \textbf{Breton} & 34.57 & 35.80 & 42.96 & 27.65 & 45.68 & 39.36 & 34.81 & 37.78 & 35.06 \\
        \textbf{Bulgarian} & 49.06 & 22.91 & 62.26 & 34.23 & 64.42 & 56.49 & 54.45 & 61.99 & 57.68 \\
        \textbf{Arabic} & 49.26 & 43.35 & 62.07 & 32.02 & 65.02 & 68.47 & 57.64 & 61.08 & 58.62 \\
        \textbf{Hindi} & 68.16 & 30.85 & 80.60 & 42.79 & 81.09 & 78.11 & 75.62 & 75.12 & 78.11 \\
        \textbf{Igbo} & 44.00 & 41.50 & 43.50 & 25.50 & 47.00 & 38.00 & 34.50 & 36.55 & 41.50 \\
        \textbf{Irish} & 53.99 & 57.67 & 64.11 & 31.29 & 65.95 & 56.13 & 55.52 & 57.98 & 57.06 \\
        \textbf{Japanese} & 50.74 & 49.26 & 53.20 & 29.06 & 53.69 & 59.11 & 58.62 & 58.62 & 57.14 \\
        \textbf{Kinyarwanda} & 35.32 & 34.47 & 53.62 & 28.51 & 53.19 & 40.43 & 32.77 & 38.30 & 40.43 \\
        \textbf{Korean} & 59.66 & 73.45 & 76.90 & 42.76 & 77.93 & 80.00 & 77.24 & 77.59 & 74.14 \\
        \textbf{Marathi} & 48.02 & 31.19 & 74.75 & 34.16 & 77.23 & 66.17 & 63.86 & 61.39 & 68.81 \\
        \textbf{Mongolian} & 39.42 & 39.74 & 50.32 & 23.40 & 50.64 & 36.01 & 44.23 & 39.10 & 47.76 \\
        \textbf{Norwegian} & 54.52 & 64.21 & 66.89 & 38.80 & 69.57 & 66.22 & 61.20 & 68.56 & 69.90 \\
        \textbf{Oromo} & 34.11 & 35.51 & 45.79 & 24.77 & 47.20 & 36.45 & 38.32 & 35.05 & 42.06 \\
        \textbf{Portuguese} & 57.75 & 73.59 & 80.63 & 41.20 & 80.63 & 78.01 & 76.06 & 76.76 & 77.46 \\
        \textbf{Romanian} & 58.94 & 67.88 & 75.17 & 40.73 & 76.49 & 74.17 & 70.20 & 75.83 & 70.20 \\
        \textbf{Russian} & 66.50 & 37.00 & 80.50 & 45.50 & 81.50 & 80.00 & 76.50 & 79.00 & 84.00 \\
        \textbf{Sinhala} & 48.00 & 28.44 & 65.33 & 29.78 & 65.78 & 39.56 & 43.56 & 45.50 & 45.78 \\
        \textbf{Spanish} & 57.02 & 69.13 & 69.34 & 37.61 & 69.92 & 73.25 & 68.20 & 71.44 & 71.55 \\
        \textbf{Swahili} & 53.11 & 60.07 & 76.19 & 36.26 & 80.95 & 66.18 & 60.07 & 55.31 & 67.77 \\
        \textbf{Telugu} & 55.50 & 32.50 & 71.50 & 32.00 & 74.50 & 57.79 & 61.00 & 58.50 & 57.00 \\
        \textbf{Urdu} & 55.98 & 48.02 & 73.39 & 36.93 & 75.69 & 66.69 & 66.99 & 64.92 & 69.50 \\
        \midrule
        \textbf{Other Avg.} & 50.81 & 45.99 & 65.39 & 33.68 & 67.06 & 58.70 & 57.32 & 58.01 & 60.25 \\
        \midrule
        \textbf{Global Avg.} & 51.53 & 49.01 & 65.69 & 31.94 & 67.59 & 60.18 & 58.83 & 59.08 & 60.86 \\
        \bottomrule
        \end{tabular}
    }
    \caption{Comparison of our best model (SEA-Gemma-3 10\%) againsts other competitor models and the regionally-adapted Gemma-3 (SEA-Gemma-3) model on CVQA.}
    \label{tab:cvqa-competitor}
\end{table*}

\begin{table}[!t]
    \centering
    \resizebox{\linewidth}{!}{
        \begin{tabular}{lccccc}
            \toprule
             & \textbf{Gemma-3} & \textbf{SEA-Gemma-3} & \textbf{SEA-Gemma-3} & \textbf{Aya-Vision} & \textbf{Qwen-3 VL} \\
            \textbf{Region} & \textbf{(27B)} & \textbf{(27B)} & \textbf{10\% (27B)} & \textbf{(32B)} & \textbf{(32B)} \\
             \midrule
            \textbf{Cambodia} & 50.7 & 40.5 & 52.3 & \textbf{56.25 }& 50.99 \\
            \textbf{Indonesia} & 43.5 & 49.3 & \textbf{69.0} & 66.09 & 66.09 \\
            \textbf{Laos} & 0.0 & 50.0 & \textbf{68.1} & 45.83 & 63.89 \\
            \textbf{Malaysia} & 37.5 & 44.4 & 62.4 & \textbf{69.31} & 60.32 \\
            \textbf{Philippines} & 63.4 & 46.6 & 58.8 & \textbf{63.40} & 56.21 \\
            \textbf{Singapore} & 56.3 & 46.9 & 68.8 & \textbf{71.88} & 65.63 \\
            \textbf{Thailand} & 51.6 & 39.1 & 52.2 & 49.46 & \textbf{52.72} \\
            \textbf{Vietnam} & 18.9 & 18.5 & 57.5 & \textbf{63.26} & 53.04 \\
            \midrule
            \textbf{SEA Avg.} & 40.2 & 41.9 & \textbf{61.1} & 60.68 & 58.61 \\
            \bottomrule
        \end{tabular}
    }
    \caption{Comparison of our best model (SEA-Gemma-3 10\%) againsts other competitor models and the regionally-adapted Gemma-3 (SEA-Gemma-3) model on SEA-VQA.}
    \label{tab:sea-vqa-competitor}
\end{table}

\section{Results Breakdown}
\label{sec:detailed-results}

We provide the results breakdown in Table~\ref{tab:model_performance_seavqa}, Table~\ref{tab:model_performance_cvqa}, Table~\ref{tab:model_performance_wc}, and Table~\ref{tab:data_curation} for SEA-VLM; Table~\ref{tab:detailed_imagegen_scores} for SEA-ImageGen, and Table~\ref{tab:detailed_siglip-seavqa} for SEA-VL Embedding.

\begin{table}[!h]
\centering
\resizebox{\linewidth}{!}{
    \begin{tabular}{lccccccccc}
    \hline
    & \textbf{Avg.} & \multicolumn{8}{c}{\textbf{Country}} \\
    \cmidrule(lr){3-10}
    \multicolumn{1}{c}{\textbf{Model}} & \textbf{Score} & \textbf{Camb.} & \textbf{Indo.} & \textbf{Laos} & \textbf{Mala.} & \textbf{Phil.} & \textbf{Sing.} & \textbf{Thai.} & \textbf{Viet.} \\
    \hline
    \textbf{Google Gemma-3} & 41.0 & 50.7 & 43.5 & 0.0 & 37.5 & \textbf{63.4} & 56.3 & {51.6} & 18.9 \\
    \midrule
    \textbf{SEA-Gemma-3 5\%} & {61.2} & {50.3} & {68.8} & {66.7} & \textbf{62.4} & 59.5 & {65.6} & 51.1 & \textbf{57.8} \\
    \textbf{SEA-Gemma-3 10\%} & \textbf{61.7} & \textbf{52.3} & \textbf{69.0} & \textbf{68.1} & \textbf{62.4} & 58.8 & \textbf{68.8} & 52.2 & {57.5} \\
    \textbf{SEA-Gemma-3 50\%} & 59.5 & 52.0 & {67.3} & 62.5 & 57.1 & 56.9 & 53.1 & \textbf{52.7} & {55.0} \\    
    \textbf{SEA-Gemma-3 70\%} & 54.0 & 51.3 & {67.3} & 63.9 & 56.6 & 58.2 & 53.1 & {51.6} & {20.5} \\
    \midrule
    \textbf{SEA-Gemma-3 (w/o merge)} & 41.9 & 40.5 & 49.3 & 50.0 & 44.4 & 46.6 & 46.9 & 39.1 & 18.5 \\
    \hline
    \end{tabular}
}
\caption{SEA-VLM Zero-shot performance on SEA-VQA. The result is broken down per country. ``Camb.'' = Cambodia, ``Indo.'' = Indonesia, ``Mala.'' = Malaysia, `Phil.' = Philippines, ``Sing.''=Singapore,  ``Thai.'' = Thailand, ``Viet.'' = Vietnam}
\label{tab:model_performance_seavqa}
\end{table}

\begin{table}[!h]
\centering
\resizebox{\linewidth}{!}{
    \begin{tabular}{lcccccc}
    \toprule
     & \textbf{Overall} & \textbf{SEA} & \multicolumn{4}{c}{\textbf{By Prompt Type}} \\
    \cmidrule{4-7}
    \textbf{Model} & \textbf{Score} & \textbf{Avg.} & \textbf{Task 1 (NC)} & \textbf{Task 2} & \textbf{Task 1 (C)} & \textbf{Task 1 (Adv.)} \\
    \midrule
    \textbf{Google Gemma-3} & 59.8 & {60.1} & \textbf{64.8} & \textbf{48.3} & {78.5} & {47.5} \\
    \midrule
    \textbf{SEA-Gemma-3 5\%} & \textbf{60.0} & \textbf{60.3} & 64.5 & 48.1 & \textbf{79.2} & \textbf{48.1} \\
    \textbf{SEA-Gemma-3 10\%} & \textbf{60.0} & 60.2 & \textbf{64.5} & 48.0 & 79.1 & \textbf{48.1} \\
    \textbf{SEA-Gemma-3 50\%} & 51.6 & 51.4 & 53.7 & 42.0 & 67.0 & 43.6 \\
    \textbf{SEA-Gemma-3 70\%} & 51.9 & 52.6 & 53.5 & 43.5 & 67.1 & 43.4 \\
    \midrule
    \textbf{SEA-Gemma-3 (w/o merge)} & 48.5 & 48.6 & 49.3 & 43.5 & 61.9 & 39.4 \\
    \bottomrule
    \end{tabular}
}
\caption{SEA-VLM Zero-shot performance on World Cuisine. (NC) = No Context, (C) = Context, (Adv.) = Adversarial}
\label{tab:model_performance_wc}
\end{table}

\begin{table}[!h]
\centering
\resizebox{\linewidth}{!}{
    \begin{tabular}{lcccc}
\hline
\multicolumn{1}{c}{\multirow{2}{*}{\textbf{Model}}} & \multicolumn{2}{c}{\textbf{Overall SEA}}               & \multicolumn{2}{c}{\textbf{Prompt Type}}                    \\ \cline{4-5} 
\multicolumn{1}{c}{}                                & \multicolumn{1}{c}{\textbf{Score}} & \textbf{Avg.} & \multicolumn{1}{c}{\textbf{English}} & \textbf{Local Lang.} \\ \hline
\textbf{Gemma-3}                                                & \multicolumn{1}{c}{67.2}           & 67.83         & \multicolumn{1}{c}{67.96}            & 66.44                \\ \midrule
\textbf{SEA-Gemma-3 5\%}                                        & \multicolumn{1}{c}{68.72}          & \textbf{69.45}         & \multicolumn{1}{c}{69.39}            & 68.05            \\
\textbf{SEA-Gemma-3 10\%}                                        & \multicolumn{1}{c}{\textbf{68.76}}          & \textbf{69.45}         & \multicolumn{1}{c}{\textbf{69.42}}            & \textbf{68.09}                \\ 
\textbf{SEA-Gemma-3 50\%}                                        & \multicolumn{1}{c}{61.84}          & 62.62         & \multicolumn{1}{c}{62.9}             & 60.79                \\
\textbf{SEA-Gemma-3 70\%}                                        & \multicolumn{1}{c}{60.58}          & 61.27         & \multicolumn{1}{c}{61.51}            & 59.66                \\ 
\midrule
\textbf{SEA-Gemma-3 (w/o merge)}                               & \multicolumn{1}{c}{35.58}          & 36.16         & \multicolumn{1}{c}{38.0}             & 33.16                \\ \bottomrule
\end{tabular}
}
\caption{SEA-VLM Zero-shot performance on CVQA.}
\label{tab:model_performance_cvqa}
\end{table}

\begin{table}[!h]
\centering
\resizebox{\linewidth}{!}{
    \begin{tabular}{lcccccccccccc}
    \toprule
     & \multicolumn{5}{c}{\textbf{Attribute}} & \multicolumn{3}{c}{\textbf{Entity}} & \multicolumn{2}{c}{\textbf{Other}} & \multicolumn{2}{c}{\textbf{Relation}} \\
    \cmidrule(lr){2-6} \cmidrule(lr){7-9} \cmidrule(lr){10-11}  \cmidrule(lr){12-13}
     \multicolumn{1}{c}{\textbf{Model}}  & \scriptsize{\textbf{color}} & \scriptsize{\textbf{other}} & \scriptsize{\textbf{shape}} & \scriptsize{\textbf{size}} & \scriptsize{\textbf{texture}} & \scriptsize{\textbf{part}} & \scriptsize{\textbf{state}} & \scriptsize{\textbf{whole}} & \scriptsize{\textbf{count}} & \scriptsize{\textbf{text}} & \scriptsize{\textbf{non-spatial}} & \scriptsize{\textbf{spatial}} \\
    \midrule
    \textbf{StabilityAI SDXL} & 80.88 & 79.49 & 69.87 & 67.36 & 79.98 & 78.52 & 74.23 & 82.71 & 56.00 & 76.00 & 81.13 & 86.73  \\
    \midrule
    \textbf{SEA-SDXL 25\%} & 82.34 & 81.49 & 75.11 & 63.22 & 80.82 & 82.81 & 73.39 & 83.45 & 59.50 & 70.00 & 77.36 & 86.93  \\
    \textbf{SEA-SDXL 50\%} & 82.39 & 80.49 & 73.80 & 66.53 & 82.52 & 80.08 & 74.13 & 82.67 & 58.50 & 66.00 & 76.73 & 87.80  \\
    \textbf{SEA-SDXL 75\%} & 81.59 & 79.49 & 72.49 & 67.36 & 81.19 & 79.49 & 73.71 & 83.09 & 61.50 & 70.00 & 81.76 & 86.23 \\   
    \midrule
    \textbf{SEA-SDXL 100\%} & 82.14 & 81.26 & 77.73 & 64.46 & 81.73 & 79.69 & 74.45 & 83.20 & 62.00 & 78.00 & 84.91 & 87.06 \\
    \bottomrule
    \end{tabular}
}
\caption{Detailed per aspect breakdown (Attribute, Entity, Other, and Relation) of the DPGBench evaluation.}
\label{tab:detailed_imagegen_scores}
\end{table}

\begin{table}[!h]
\centering
\resizebox{\linewidth}{!}{
    \begin{tabular}{lccccccccc}
    \toprule
    \textbf{Model} & \multicolumn{9}{c}{\textbf{SEAVQA Zero-Shot Performance}} \\
    \cmidrule(lr){2-10}
     & \scriptsize{\textbf{Overall}} & \scriptsize{\textbf{Cambodia}} & \scriptsize{\textbf{Indonesia}} & \scriptsize{\textbf{Laos}} & \scriptsize{\textbf{Malaysia}} & \scriptsize{\textbf{Phillipines}} & \scriptsize{\textbf{Singapore}} & \scriptsize{\textbf{Thailand}} & \scriptsize{\textbf{Vietnam}} \\
    \midrule
    \textbf{Google SigLIP2} & 25.81 & 28.9 & 25.0 & 22.2 & 24.3 & 28.8 & 21.9 & 25.5 & 25.6 \\
    \midrule
    \textbf{SEA-SigLIP2 25\%} & 26.36 & 27.6 & 25.4 & 29.2 & 29.1 & 28.8 & 28.1 & 26.1 & 24.0 \\
    \textbf{SEA-SigLIP2 50\%} & 28.06 & 31.3 & 27.0 & 22.2 & 29.6 & 27.5 & 21.9 & 31.0 & 27.2 \\
    \textbf{SEA-SigLIP2 75\%} & 29.66 & 28.6 & 30.1 & 19.4 & 32.3 & 34.0 & 25.0 & 29.9 & 28.8 \\
    \midrule
    \textbf{SEA-SigLIP2 100\%} & 28.96 & 28.3 & 28.6 & 29.2 & 26.5 & 32.7 & 31.3 & 30.4 & 29.1 \\
    \bottomrule
    \end{tabular}
}
\caption{Detailed per country zero-shot performance of the SEAVQA evaluation.}
\label{tab:detailed_siglip-seavqa}
\end{table}

\begin{table}[h!]
\centering
\resizebox{\linewidth}{!}{
    \begin{tabular}{lccccccccc}
    \hline
    & \textbf{Avg.} & \multicolumn{8}{c}{\textbf{Country}} \\
    \cmidrule(lr){3-10}
    \multicolumn{1}{c}{\textbf{Model}} & \textbf{Score} & \textbf{Camb.} & \textbf{Indo.} & \textbf{Laos} & \textbf{Mala.} & \textbf{Phil.} & \textbf{Sing.} & \textbf{Thai.} & \textbf{Viet.} \\
    \hline
    \midrule
    \textbf{SEA-Mammoth 50k} & 13.0 & 9.9 & 18.8 & 11.1 & 11.1 & 11.8 & 18.8 & 14.67 & 2.88 \\
    \textbf{SEA-Mammoth 250k} & 39.6 & 13.8 & \textbf{52.9} & \textbf{52.8} & 42.9 & \textbf{47.1} & 46.9 & \textbf{44.0} & \textbf{20.8} \\ 
    \textbf{SEA-Mammoth 250k + CG OE} & \textbf{41.9} & \textbf{40.5} & 49.3 & 50.0 & \textbf{44.4} & 46.6 & 46.9 & 39.1 & 18.5 \\ 
    \textbf{SEA-Mammoth 250k + CG MC} & 21.6 & 26.6 & 20.9 & 26.4 & 16.4 & 24.8 & 28.1 & 28.3 & 14.1 \\ 
    \textbf{SEA-Mammoth 250k + WC} & 30.4 & 10.2 & 39.9 & 34.7 & 37.6 & 34.0 & \textbf{50.0} & 32.6 & 16.9 \\ 
    \hline
    \end{tabular}
}
\caption{SEA-VLM Zero-shot performance on SEA-VQA. The result is broken down per country. ``Camb.'' = Cambodia,  ``IndO.'' = Indonesia, ``Mala.'' = Malaysia, `Phil.' = Phillipines, ``Sing.''=Singapore,  ``Thai.'' = Thailand, ``Viet.'' = Vietnam. ``CG'' denotes CulturalGround, ``WC'' denotes WorldCuisine, while ``OE'' and ``MC'' denote open-ended and multiple-choice, respectively.}
\label{tab:data_curation}
\end{table}

\section{Sample Outputs of SEA-VLM and SEA-ImageGen}
\label{sec:sample-output}

\begin{figure}[!t]
    \centering
    \includegraphics[width=0.9\linewidth, trim={0 1.5cm 0 0.5cm}, clip]{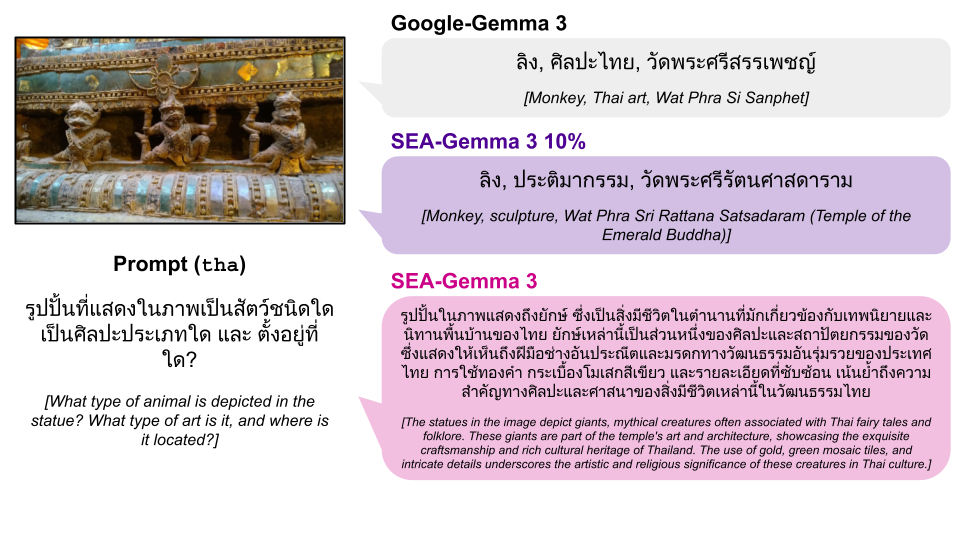}
    \includegraphics[width=0.9\linewidth, trim={0 2.5cm 0 0.5cm}, clip]{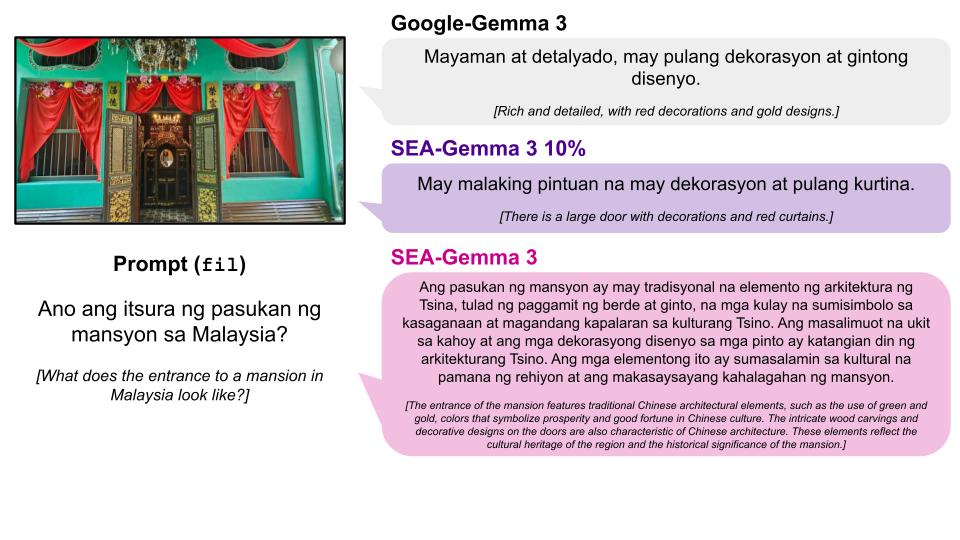}
    \includegraphics[width=0.9\linewidth, trim={0 0.2cm 0 0.5cm}, clip]{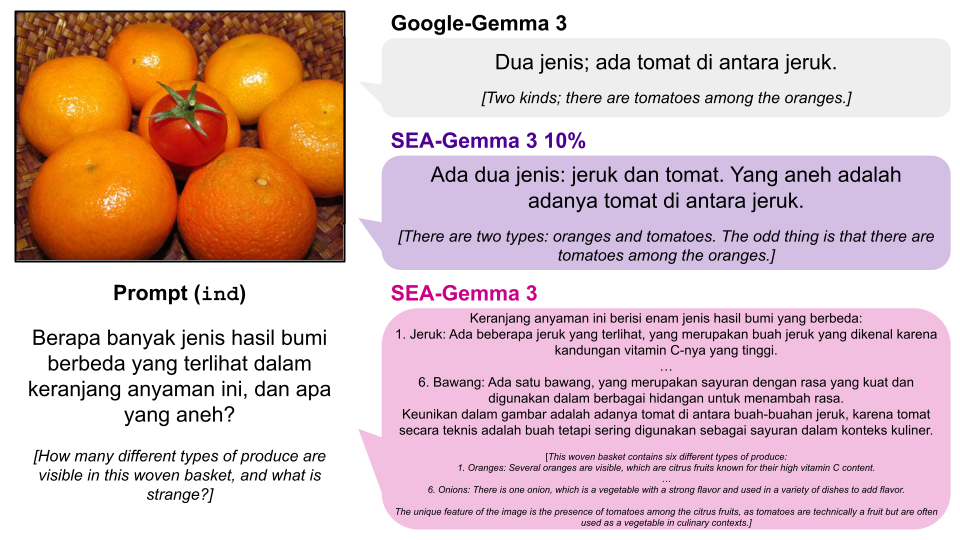}
    \caption{Generated responses using different model archetypes. From left-to-right: global model (Gemma-3), our regional model (SEA-Gemma-3), our merged model (SEA-Gemma-3 10\%) along with the prompts. Our model produces the most correct caption among others, while retaining the image naturalness and overall quality of the original Gemma-3.}
    \label{fig:vlm-samples}
\end{figure}

\begin{figure}[!t]
    \centering
    \includegraphics[width=\linewidth]{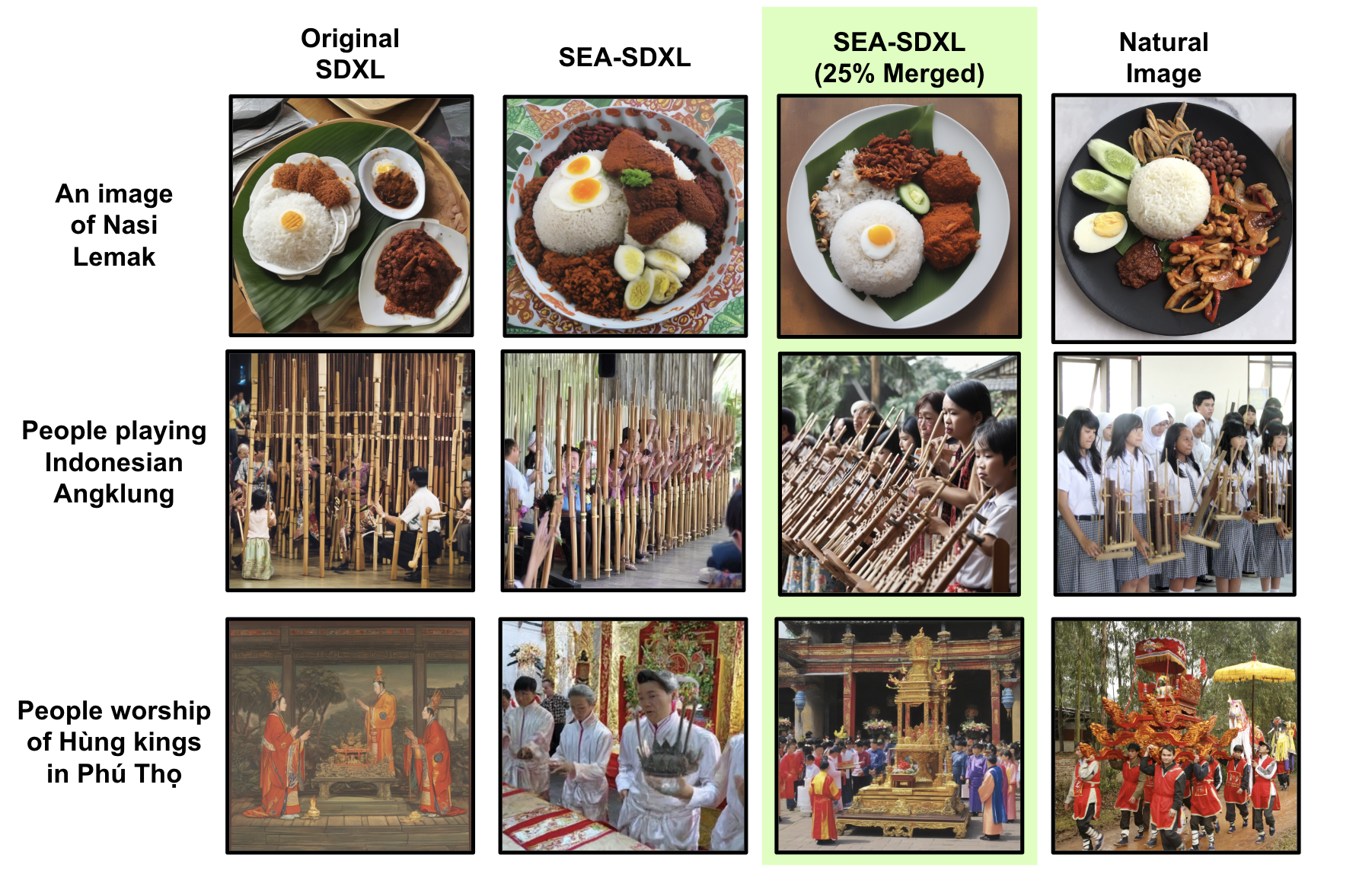}
    \caption{Generated image using different model archetypes. From left-to-right: global model (SDXL), our regional model (SEA-SDXL), our merged model (SEA-SDXL 25\%), and reference of natural images.}
    \label{fig:image-gen-samples}
\end{figure}

We provide samples generated by our best SEA-VLM and SEA-ImageGen compared to the baselines on Figure~\ref{fig:vlm-samples}  and Figure~\ref{fig:image-gen-samples}, respectively. As shown in Figure~\ref{fig:image-gen-samples}, the SEA-SDXL 25\% model produces the most accurate representation among others, while retaining the image naturalness and overall quality of the original SDXL. For example, while the baselines also capture the component of nasi lemak such as nasi putih (white rice) and telur (egg), the SEA-SDXL 25\% model incorporates more detailed factors including timun (cucumber), kacang (peanut),  ikan bilis (anchovy). Similarly for cultural and tradition categories, SEA-SDXL 25\% captures a more accurate illustration of playing angklung and Hung king festival in Phu Tho, Vietnam.

\section{Globalization Index by Region by Year}
\label{sec:kof-gi}

We provide the detailed per year per region globalization index which we used for deciding the globalization factor $\alpha$ in GG-EZ in Table~\ref{tab:globalization_index}.

\begin{table*}[h!]
\centering
\resizebox{0.8\linewidth}{!}{
    \begin{tabular}{
        >{\centering\arraybackslash}m{1.25cm}
        >{\centering\arraybackslash}m{1cm}
        >{\centering\arraybackslash}m{1cm}
        >{\centering\arraybackslash}m{1cm}
        >{\centering\arraybackslash}m{1cm}
        >{\centering\arraybackslash}m{1cm}
        >{\centering\arraybackslash}m{1cm}
        >{\centering\arraybackslash}m{1cm}
        >{\centering\arraybackslash}m{1cm}
        >{\centering\arraybackslash}m{1cm}
    }
    \toprule
    \textbf{Year} & \textbf{EAP} & \textbf{ECA} & \textbf{LAC} & \textbf{MENA} & \textbf{NA} & \textbf{SA} & \textbf{SEA} & \textbf{SSA} & \textbf{World} \\
    \midrule
    \textbf{1993} & 49.32 & 58.82 & 51.20 & 57.00 & 68.59 & 28.05 & 36.04 & 34.37 & 48.86 \\
    \textbf{1994} & 51.07 & 59.15 & 51.00 & 56.85 & 68.58 & 28.25 & 36.40 & 34.26 & 49.15 \\
    \textbf{1995} & 51.40 & 59.72 & 51.67 & 56.62 & 68.66 & 28.39 & 37.34 & 34.13 & 49.42 \\
    \textbf{1996} & 52.02 & 60.23 & 51.67 & 56.74 & 69.43 & 28.37 & 37.89 & 33.99 & 49.64 \\
    \textbf{1997} & 51.68 & 60.43 & 51.75 & 56.79 & 69.48 & 28.35 & 37.76 & 33.43 & 49.53 \\
    \textbf{1998} & 51.63 & 60.71 & 51.86 & 56.89 & 69.83 & 28.33 & 37.70 & 33.55 & 49.66 \\
    \textbf{1999} & 52.00 & 61.00 & 52.20 & 56.94 & 70.64 & 28.26 & 38.59 & 33.45 & 49.85 \\
    \textbf{2000} & 52.48 & 61.65 & 53.33 & 57.03 & 71.09 & 28.94 & 39.23 & 33.63 & 50.38 \\
    \textbf{2001} & 52.61 & 62.06 & 53.88 & 57.43 & 71.74 & 28.96 & 39.12 & 33.80 & 50.70 \\
    \textbf{2002} & 52.63 & 62.83 & 54.53 & 57.66 & 72.50 & 30.16 & 40.06 & 33.92 & 51.15 \\
    \textbf{2003} & 52.75 & 63.55 & 55.32 & 58.04 & 73.02 & 30.84 & 39.76 & 34.82 & 51.79 \\
    \textbf{2004} & 53.41 & 64.48 & 55.88 & 58.42 & 73.52 & 31.06 & 40.63 & 34.95 & 52.33 \\
    \textbf{2005} & 53.93 & 65.47 & 56.19 & 58.48 & 74.17 & 32.29 & 41.61 & 35.24 & 52.86 \\
    \textbf{2006} & 54.32 & 66.19 & 56.86 & 59.34 & 74.39 & 32.87 & 41.96 & 35.70 & 53.46 \\
    \textbf{2007} & 55.16 & 67.67 & 58.04 & 59.78 & 75.24 & 33.99 & 42.90 & 36.40 & 54.46 \\
    \textbf{2008} & 55.55 & 68.76 & 58.41 & 60.78 & 75.57 & 36.58 & 43.59 & 36.72 & 55.17 \\
    \textbf{2009} & 55.70 & 68.42 & 58.59 & 61.54 & 75.41 & 37.98 & 44.27 & 37.16 & 55.38 \\
    \textbf{2010} & 55.94 & 68.92 & 58.37 & 62.05 & 75.87 & 39.49 & 43.19 & 37.37 & 55.68 \\
    \textbf{2011} & 56.40 & 70.04 & 58.50 & 62.51 & 76.43 & 40.59 & 44.17 & 38.87 & 56.53 \\
    \textbf{2012} & 56.55 & 70.52 & 58.75 & 63.24 & 76.64 & 42.27 & 44.50 & 38.92 & 56.88 \\
    \textbf{2013} & 56.50 & 70.93 & 59.06 & 63.55 & 77.25 & 42.87 & 45.33 & 39.14 & 57.16 \\
    \textbf{2014} & 56.77 & 71.07 & 59.00 & 63.44 & 77.80 & 43.90 & 46.15 & 39.54 & 57.36 \\
    \textbf{2015} & 57.13 & 70.32 & 58.76 & 62.50 & 77.86 & 44.75 & 45.79 & 38.99 & 56.98 \\
    \textbf{2016} & 57.09 & 70.13 & 58.97 & 62.44 & 78.44 & 44.38 & 45.58 & 37.91 & 56.70 \\
    \textbf{2017} & 57.04 & 70.88 & 59.00 & 60.91 & 78.81 & 44.14 & 45.12 & 37.71 & 56.75 \\
    \textbf{2018} & 57.34 & 71.10 & 58.85 & 60.79 & 78.82 & 43.25 & 46.21 & 37.15 & 56.65 \\
    \textbf{2019} & 56.36 & 70.84 & 58.51 & 60.92 & 78.95 & 40.53 & 44.29 & 36.54 & 56.12 \\
    \textbf{2020} & 54.54 & 69.16 & 56.97 & 59.02 & 77.72 & 38.80 & 41.33 & 34.97 & 54.45 \\
    \textbf{2021} & 53.94 & 69.60 & 57.23 & 59.66 & 78.02 & 37.97 & 39.87 & 34.86 & 54.54 \\
    \textbf{2022} & 54.96 & 70.78 & 58.30 & 60.00 & 79.04 & 39.20 & 42.11 & 35.36 & 55.42 \\
    \textbf{2023} & 55.74 & 71.03 & 58.91 & 60.13 & 79.36 & 40.06 & 43.40 & 35.46 & 55.79 \\
    \bottomrule
    \end{tabular}
}
\caption{Globalization index by region by year from 1993 to 2023. ``EAP'' = East Asia and Pacific , `ECA'' = Europe and Central Asia,``LAC'' = Latin America and Caribbean, ``MENA'' = Middle East and North Africa, ``NA'' = North America, ``SA'' = South Asia, ``SEA'' = Southeast Asian, ``SSA'' = Sub-Saharan Africa.}
\label{tab:globalization_index}
\end{table*}

\section{Compute Resources}
\label{sec:compute-resources}

For all of our experiments, we a single H100 compute node with 8 H100 GPUs. The biggest computes come from running CPT and SFT  of VLM models which takes use of the full compute for several days per experiment. while other experiments, i.e., SEA-ImageGen and SEA-VL Embed, uses only 2 H100 GPUs for each experiment.

\end{document}